\documentclass[11pt, twocolumn]{article}

\usepackage{microtype}
\usepackage{graphicx}
\usepackage{booktabs} 

\usepackage{hyperref}

\usepackage{amsmath}
\usepackage{amssymb}
\usepackage{mathtools}
\usepackage{amsthm}

\usepackage{graphicx}%
\usepackage{multirow}%
\usepackage{hyperref}
\usepackage{amsmath,amssymb,amsfonts}%
\usepackage{amsthm}%
\usepackage{tabularx}
\usepackage{mathabx}
\usepackage{mathrsfs}%
\usepackage[title]{appendix}%
\usepackage{xcolor}%
\usepackage{textcomp}%
\usepackage{manyfoot}%
\usepackage{booktabs}%
\usepackage{listings}%
\usepackage{comment}

\usepackage{caption}
\usepackage{subcaption} 

\usepackage[utf8]{inputenc}
\usepackage{bbm}
\usepackage{mathtools}
\usepackage{changepage}
\usepackage{xspace}
\usepackage[backend=bibtex]{biblatex}
\bibliography{references}

\DeclareMathOperator*{\argmin}{arg\,min}

\newcommand{\rulesep}{\unskip\ \vline height 10.5ex\ }

\theoremstyle{plain}
\newtheorem{theorem}{Theorem}[section]
\newtheorem{proposition}[theorem]{Proposition}
\newtheorem{lemma}[theorem]{Lemma}
\newtheorem{corollary}[theorem]{Corollary}
\theoremstyle{definition}

\theoremstyle{remark}

\newcommand{\tr}{\ensuremath{\text{Tr}}\xspace}

\newcommand{\I}{\ensuremath{\mathbf{I}}\xspace}
\newcommand{\J}{\ensuremath{\mathbf{J}}\xspace}
\newcommand{\Lmat}{\ensuremath{\mathbf{L}}\xspace}
\newcommand{\Omat}{\ensuremath{\mathbf{O}}\xspace}
\newcommand{\C}{\ensuremath{\mathbf{C}}\xspace}

\newcommand{\A}{\ensuremath{\mathbf{A}}\xspace}
\newcommand{\B}{\ensuremath{\mathbf{B}}\xspace}

\newcommand{\M}{\ensuremath{\mathbf{M}}\xspace}

\newcommand{\La}{\ensuremath{\mathbf{\Lambda}}\xspace}

\newcommand{\K}{\ensuremath{\mathbf{K}}\xspace}
\newcommand{\Kx}{\ensuremath{\mathbf{K}_X}\xspace}
\newcommand{\Ky}{\ensuremath{\mathbf{K}_Y}\xspace}

\newcommand{\Gx}{\ensuremath{\mathbf{G}_X}\xspace}
\newcommand{\Gy}{\ensuremath{\mathbf{G}_Y}\xspace}
\newcommand{\Dx}{\ensuremath{\mathbf{D}_X}\xspace}
\newcommand{\Dy}{\ensuremath{\mathbf{D}_Y}\xspace}

\newcommand{\Ux}{\ensuremath{\mathbf{U}_X}\xspace}
\newcommand{\Uy}{\ensuremath{\mathbf{U}_Y}\xspace}
\newcommand{\Sx}{\ensuremath{\mathbf{\Sigma}_X}\xspace}
\newcommand{\Sy}{\ensuremath{\mathbf{\Sigma}_Y}\xspace}
\newcommand{\Vx}{\ensuremath{\mathbf{V}_X}\xspace}
\newcommand{\Vy}{\ensuremath{\mathbf{V}_Y}\xspace}

\newcommand{\Uxp}{\ensuremath{\mathbf{U}_X^{+}}\xspace}
\newcommand{\Uxm}{\ensuremath{\mathbf{U}_X^{-}}\xspace}
\newcommand{\Sxp}{\ensuremath{\mathbf{\Sigma}_X^{+}}\xspace}
\newcommand{\Sxm}{\ensuremath{\mathbf{\Sigma}_X^{-}}\xspace}
\newcommand{\Vxp}{\ensuremath{\mathbf{V}_X^{+}}\xspace}
\newcommand{\Vxm}{\ensuremath{\mathbf{V}_X^{-}}\xspace}

\newcommand{\W}{\ensuremath{\mathbf{W}}\xspace}
\newcommand{\X}{\ensuremath{\mathbf{X}}\xspace}
\newcommand{\Y}{\ensuremath{\mathbf{Y}}\xspace}

\newcommand{\Rx}{\ensuremath{\mathbf{R}_X}\xspace}
\newcommand{\Ry}{\ensuremath{\mathbf{R}_Y}\xspace}

\usepackage[textsize=tiny]{todonotes}

\begin{document}

\title{Relating tSNE and UMAP to Classical Dimensionality Reduction}
\author{Andrew Draganov, Simon Dohn\\Aarhus University}

\maketitle

\begin{abstract}

    It has become standard to use gradient-based dimensionality reduction (DR) methods like tSNE and UMAP when explaining what AI models have learned. This
    makes sense -- these methods are fast, robust, and have an uncanny ability to find semantic patterns in high-dimensional data without supervision. Despite
    this, gradient-based DR methods lack the most important quality that an explainability method should possess -- themselves being explainable.  That is,
    given a UMAP output, it is currently unclear what one can say about the corresponding input.

    We work towards closing this question by relating UMAP to classical DR techniques. Specifically, we show that one can fully recover methods like
    PCA, MDS, and ISOMAP in the modern DR paradigm -- by applying attractions and repulsions onto a randomly initialized dataset. We also show that, with
    a small change, Locally Linear Embeddings (LLE) can indistinguishably reproduce UMAP outputs. This implies that UMAP's effective objective is minimized by
    this modified version of LLE (and vice versa). Given this, we discuss what must be true of UMAP emebddings and present avenues for future work.

\end{abstract}

\section{Introduction}

Although they are not thought of as such, gradient-based dimensionality reduction (DR) techniques have emerged as important methods for Explainable AI,
especially in the unsupervised learning setting. Indeed, in the absence of class labels, the natural first step in answering ``what has my AI model learned?''
is to embed its latent space into 2D and qualitatively assess the resulting structure. To this end, UMAP~\cite{umap} and its sister method
tSNE~\cite{tsne_orig} have emerged as the de-facto standard\footnote{We evidence this in Section \ref{bad_science_list} of the Appendix, where we show
a non-exhaustive list of $\sim \!\! 50$ examples where tSNE and UMAP are used to describe learned representations.} techniques. Their widespread use can be
attributed to the abundance of positive qualities that tSNE and UMAP possess -- these methods are fast, effective, and, most importantly, have a remarkable
ability to capture the relevant information in a high-dimensional dataset.

\begin{figure*}
    \begin{subfigure}[b]{0.48\textwidth}
        \begin{subfigure}[b]{0.3215\textwidth}
            \includegraphics[width=\linewidth]{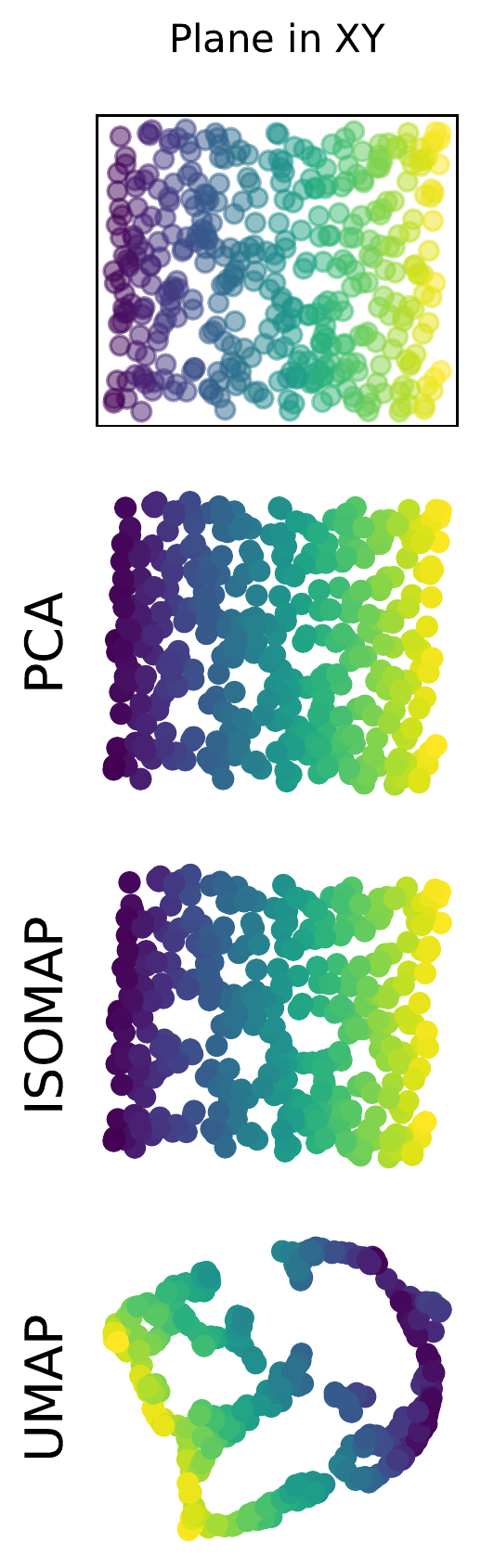}
        \end{subfigure}
        \hspace*{0.15cm}
        \begin{subfigure}[b]{0.28\textwidth}
            \includegraphics[width=\linewidth]{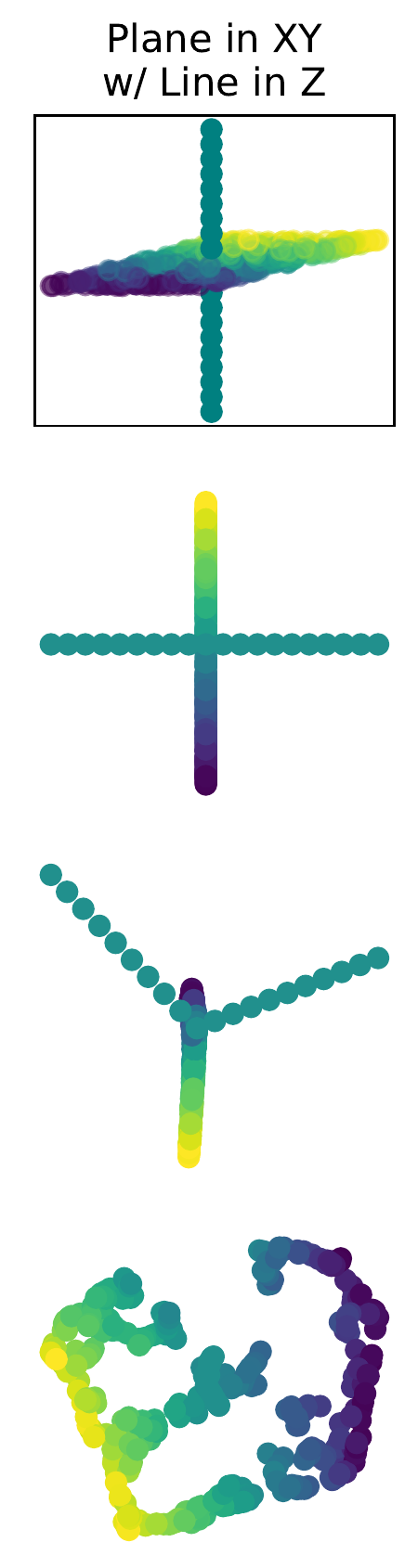}
        \end{subfigure}
        \hspace*{0.15cm}
        \begin{subfigure}[b]{0.28\textwidth}
            \includegraphics[width=\linewidth]{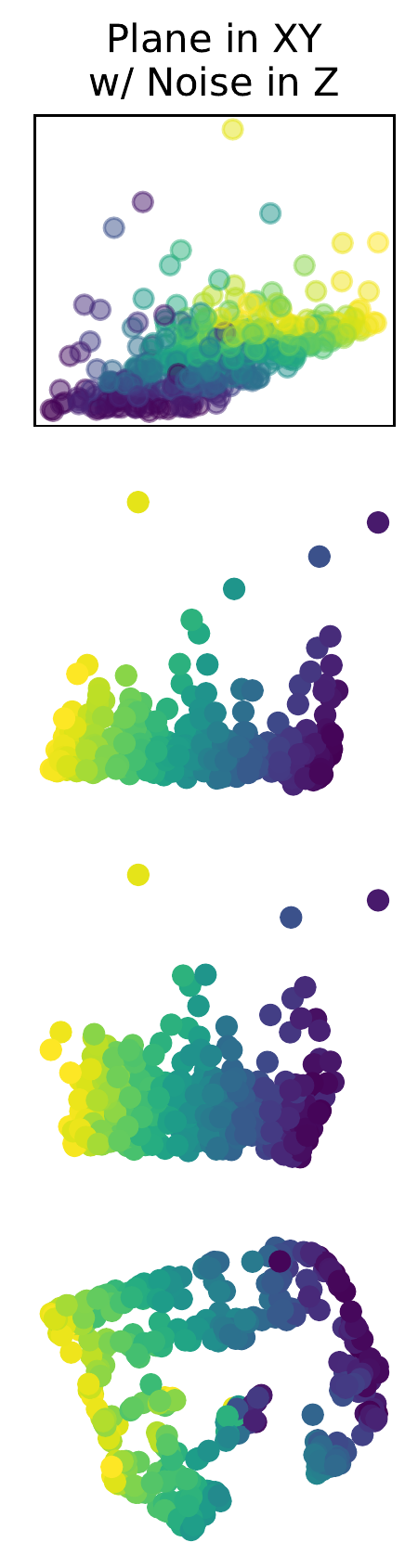}
        \end{subfigure}
        \caption{
        Examples of UMAP outputs remaining consistent as input's structure changes. The 3D datasets correspond to a 2D plane, a 2D
        plane with an orthogonal line of points, and a 2D plane with heavy-tailed (pareto w/ $\alpha=1.5$) noise in the $3^{\text{rd}}$ dimension. PCA
        and Isomap capture the change but UMAP does not.
        }
        \label{subfig:umap_doesnt_change}
    \end{subfigure}
    \hspace*{0.1cm}
    \rulesep
    \hspace*{0.1cm}
    \begin{subfigure}[b]{0.48\textwidth}
        \hspace*{0.07cm}
        \begin{subfigure}[b]{0.281\textwidth}
            \includegraphics[width=\linewidth]{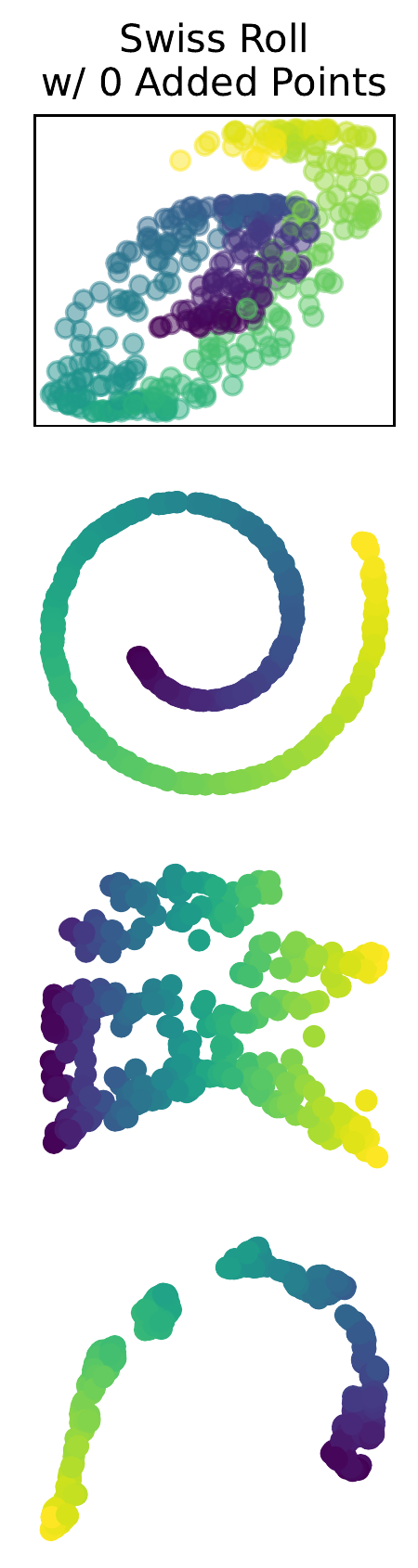}
        \end{subfigure}
        \hspace*{0.07cm}
        \rulesep
        \hspace*{0.07cm}
        \begin{subfigure}[b]{0.285\textwidth}
            \includegraphics[width=\linewidth]{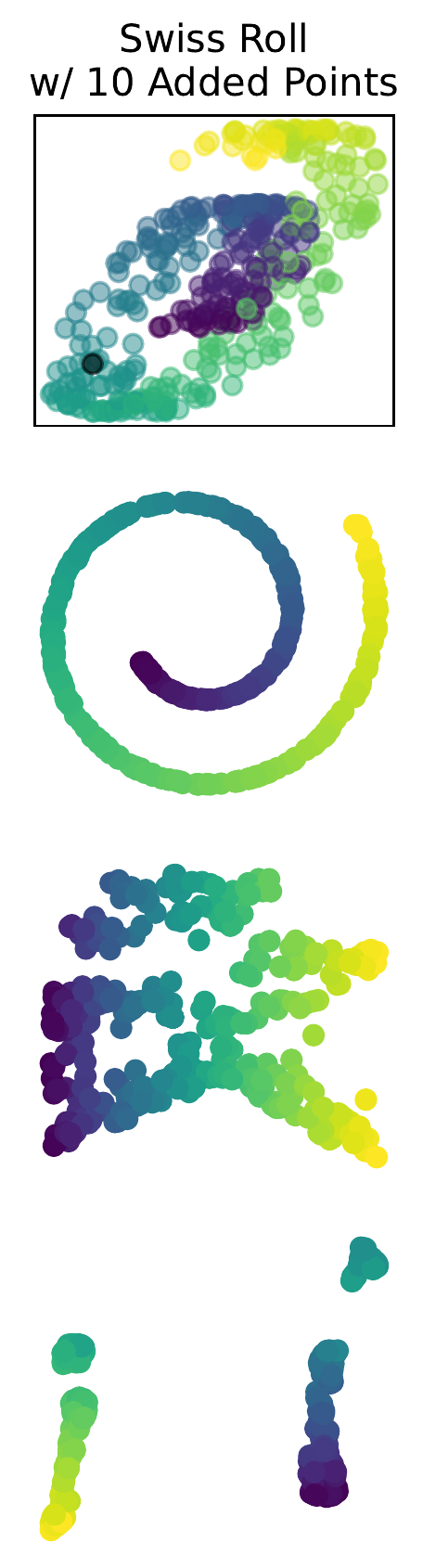}
        \end{subfigure}
        \hspace*{0.07cm}
        \rulesep
        \hspace*{0.07cm}
        \begin{subfigure}[b]{0.285\textwidth}
            \includegraphics[width=\linewidth]{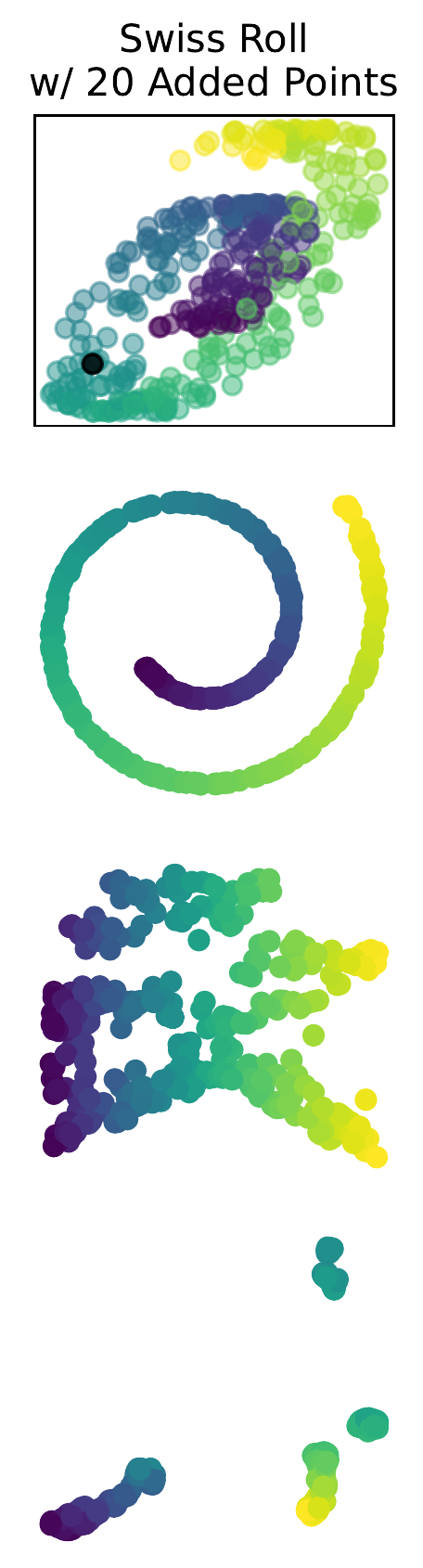}
        \end{subfigure}
        \caption{
        Examples of UMAP outputs changing as the input's structure remains consistent. The swiss rolls have $0$, $10$, and $20$ points added
        directly on the manifold, implying that there is no perturbation to the swiss roll's surface. PCA and Isomap outputs remain unchanged while UMAP outputs
        vary significantly.
        }
        \label{subfig:umap_changes}
    \end{subfigure}
    \caption{Examples of UMAP outputs behaving unintuitively: (a) they can remain stable despite the input's structure changing; (b) they can change despite
    the input's structure remaining stable.}
    \label{fig:invariances}
\end{figure*}

However, behind all of the great reasons to use gradient-based DR methods lies an inconvenient truth: it is unclear how to analyze tSNE or UMAP
embeddings~\cite{tsne_umap_init, dlp}. In the case of UMAP, this lack of clarity is due to the extensive heuristics and hyperparameters in its implementation,
which leads to misalignment between the method and its theoretical underpinning~\cite{actup}. This presents challenges for theoretically analyzing UMAP: it
took several years to determine what objective the UMAP algorithm is even optimizing in practice \cite{umap_effective_loss}. To this day, it is unclear what
convergence guarantees (if any) UMAP has. As a result, it is difficult to say what any given UMAP output actually guarantees about the high-dimensional dataset
it was fit on.  We refer to Figure \ref{fig:invariances} to see surprising settings where UMAP embeddings do and do not respond to changes in the input. Put
simply, given a UMAP embedding, little can confidently be said about the structure of the corresponding input.

Luckily, gradient-based DR techniques exist in a landscape of classical methods that do not have an explainability issue. For example, Principal Component
Analysis (PCA) maximally preserves variance from the input distribution. Thus, given a PCA embedding, one can make precise claims regarding the input's
variances along different components. Similar claims can be made for other classic DR techniques such as Locally Linear Embeddings (LLE) \cite{lle,
lle_tutorial}, Isomap~\cite{isomap}, Laplacian Eigenmaps~\cite{laplacian_eigenmaps}, and Multi-Dimensional Scaling (MDS) \cite{mds_original}, to name a few.
Importantly, each algorithm's explainability is a direct consequence of its clear objective function and convergence guarantees. Thus, given an output of
any of these algorithms, one can appropriately describe the input. 

\subsection{Our Contributions}

Our work aims to bridge this gap by showing how to get classical methods from the modern ones and vice versa.  We start by defining a framework for representing
methods such as PCA, MDS, Isomap, and LLE in the tSNE/UMAP setting. Specifically, we show that one can provably obtain PCA, MDS and Isomap embeddings by
performing attractions and repulsions between points and that this is robust to fast low-rank approximations. To our knowledge, this is the first verification
that these classical techniques are recoverable under the attraction/repulsion framework.

We then flip our perspective and arrive at a surprising conjecture: that UMAP embeddings can be reproduced using classical DR techniques. Specifically, we
define the double-kernel LLE (DK-LLE) task to be LLE with nonlinear kernels on the input and the output. We then show that one can consistently obtain UMAP
embeddings by simply minimizing the DK-LLE objective via standard gradient descent. Importantly, this avoids all of the optimizaiton heuristics present in the
UMAP algorithm and is therefore significantly simpler to formalize and study. Using these results, we provide guidance as to what UMAP embeddings should
guarantee about their inputs -- an open question that has been discussed in~\cite{umap_effective_loss, dlp, pacmap_analysis, trimap}.

\section{Preliminaries and Related Work}
\label{sec:prelim}

\subsection{Classical Methods}
\label{ssec:classical}
\paragraph{PCA, MDS and Isomap.}

Likely the most famous dimensionality reduction algorithm, PCA finds the orthogonal basis in $\mathbb{R}^{n \times d}$ that maximally preserves the variance in
the centered dataset~\cite{pca_tutorial}. Let $\X \in \mathbb{R}^{n \times D}$ be the input dataset and let $\C \X = \Ux \Sx \Vx^\top$ be the singular value
decomposition (SVD) of \X after centering\footnote{$\C = \I - \frac{1}{n}\J$ is the centering matrix with \J the all-1 matrix}. This gives us an expression for
the positive semidefinite (psd) Gram matrix $\C \Gx \C = \C \X \X^\top \C = \Ux \Sx ^2 \Ux^\top$. The principal components of $\C \X$ are defined as $\Ux \Sx$
and can therefore be found via eigen-decomposition on $\C \Gx \C$. This has a natural extension to kernel methods~\cite{kernel_pca}. Suppose that the kernel
function $k_x(x_i, x_j) = \langle \phi(x_i), \phi(x_j) \rangle$ corresponds to an inner product space. Then the kernel matrix $\K_{ij} = k(x_i, x_j)$ is a psd
Gram matrix for the space defined by $\phi$ and can be substituted in place of $\Gx$.

We now use the above setup to describe classical MDS. Rather than receiving a dataset \X as input, classical MDS accepts the $n \times n$ matrix of pairwise
distances $\Dx$ and produces the $\Y \in \mathbb{R}^{n \times d}$ such that the corresponding Euclidean distance matrix \Dy most closely approximates \Dx. Based
on the Euclidean identity that $\C \Dx \C = \frac{-1}{2} \C \Gx \C$, the traditional method for performing Classical MDS on Euclidean \X is to simply turn it
into a PCA problem. This can again be paired with kernel methods as discussed in \cite{kernel_trick_distances}. Furthermore, many DR methods such as Isomap and
PTU \cite{ptu} use classical MDS as their final step. Specifically, they find a non-Euclidean $\Dx$ matrix\footnote{In Isomap's case, this corresponds to the
distance along the $k$-nearest neighbor graph in \X} and perform MDS to find the Euclidean embedding that best represents \Dx.

\paragraph{Locally Linear Embedding (LLE).}

While PCA and MDS operate on global distances, LLE instead preserves local neighborhoods. We start with the $k$-nearest-neighbor graph on \X and define $K_i
= \{e_{i1}, \cdots e_{ik}\}$ to represent the indices of $x_i$'s $k$ nearest neighbors. We do not consider $x_i$ as its own neighbor. LLE first
finds the $\W \in \mathbb{R}^{n \times n}$ s.t. \[\sum_i ||x_i - \sum_{j \in K_i} w_{ij} x_{j}||_2^2 = ||\X - \W \X||_F^2 = ||(\I - \W)\X||_F^2\] is
minimized under the constraint $\sum_{j \in K_i} w_{ij} = 1$ for all $i$. That is, we find the weights such that each point $x_i$ is represented as a linear
combination of its neighbors. This implies that the $i$-th row of \W must be zero on the indices $l$ where $x_l$ is not $x_i$'s nearest neighbor. It can be
shown that such a \W always exists and can be found by eigendecompositions \cite{lle, lle_tutorial}.

Having found the \W that represents neighborhoods in \X, step 2 of LLE then finds the $\Y \in \mathbb{R}^{n \times d}$ such that \[\sum_i
||y_i - \sum_{j \in K_i} w_{ij} y_{j}||_2^2 = ||\Y - \W \Y||_F^2\] is minimized\footnote{Note that the \W and $K_i$ are the same as in the first step.},
subject to the constraint that the columns of \Y form an orthogonal basis, i.e. $\frac{1}{n}\Y^\top \Y = \I$.  The embedding \Y is given by the eigenvectors of
$\M = (\I - \W)^\top (\I - \W)$ that correspond to the smallest $d$ positive eigenvalues (Derivation in \ref{apx:lle_eig}).

LLE can again be combined with kernels on \X to perform more sophisticated similarity calculations~\cite{lle_tutorial}. Let \Kx be defined as in Section
\ref{ssec:classical} and let $k_y(y_i, y_j) = \langle \psi(y_i), \psi(y_j) \rangle$ be a psd kernel that defines a corresponding $\Ky \in \mathbb{R}^{n \times
n}$ with $[\Ky]_{ij} = k_y(y_i, y_j)$. Then we can define the objective function during the first step as finding the \W that minimizes $\tr \left( (\I - \W)
\Kx (\I - \W)^\top \right)$ such that $\sum_j w_{ij} = 1$ for all $i$. Similarly, the second step under a kernel function \Ky seeks the \Y such that $\tr \left(
(\I - \W) \Ky (\I - \W)^\top \right)$ is minimized.

\subsection{Gradient Dimensionality Reduction Methods}

We now switch gears from the directly solvable methods to the gradient-based ones. We will refer to the class of gradient-based dimensionality reduction
techniques that includes tSNE and UMAP as Attraction/Repulsion DR (ARDR) methods. The common theme among these is that they define notions of similarity in the
input \X and the embedding \Y. They then minimize a loss function by attracting points in \Y that should be similar and repelling points that should be
dissimilar. A non-exhaustive list of methods includes tSNE~\cite{tsne_orig, tsne_acc}, UMAP~\cite{umap}, ForceAtlas2~\cite{forceatlas2},
LargeVis~\cite{largevis} and PacMAP~\cite{pacmap}. Similar methods such as TriMAP~\cite{trimap} discuss this in the context of triplets but the schema of
gradient descent by attractions/repulsions is the same.

\subsubsection{In theory}
\label{sssec:theory}
\paragraph{ARDR methods}

Assume that we are given an input $\X \in \mathbb{R}^{n \times D}$, an embedding $\Y \in \mathbb{R}^{n \times d}$ and two non-linear psd kernel functions
$k_x(x_i, x_j) = \langle \phi(x_i), \phi(x_j) \rangle $ and $k_y(y_i, y_j) = \langle \psi(y_i), \psi(y_j) \rangle$ such that $0 \leq k_x, k_y \leq 1$.  These
then define psd matrices $\Kx, \Ky \in \mathbb{R}^{n \times n}$, where each $(i, j)$-th entry represents the set of per-point similarities in \X and \Y
respectively. Given this setup, the goal of gradient-based DR methods is to find the embedding \Y such that \Ky is maximally similar to \Kx with respect to some
matrix-wise loss function.

For the remainder of this paper, the reader can assume that $k_x$ is the exponential RBF kernel and $k_y$ is the quadratic Cauchy kernel\footnote{However, we
note that the upcoming generalizations only require that $k_y$ be a function of the squared Euclidean distance. We will write $k_y(||y_i - y_j||_2^2)$ when
emphasizing this point.}. Since the kernels are chosen such that $k_x$ and $k_y$ are necessarily in $[0, 1]$, the matrices can be treated as probability
distributions\footnote{Either as a matrix of $n^2$ Bernoulli variables in UMAP or as a single prob. distribution in tSNE (since tSNE's matrix sums to $1$).}.
Thus, the question of how well \Ky represents \Kx is traditionally quantified using the KL-divergence $KL(\Kx || \Ky)$. We refer to \cite{actup} for derivations
of the tSNE/UMAP loss functions as a sum of attraction and repulsion terms. This leads to gradients that, in their most general form, can be written as:

\vspace*{-0.5cm}
\begin{align}
    \label{eq:gradient_form}
    \nabla_{y_i} &= -c \sum_j \left( \mathcal{A}_{ij} - \mathcal{R}_{ij} \right) \cdot \dfrac{d\; k_y(||y_i - y_j||_2^2)}{d\; ||y_i - y_j||_2^2} \cdot (y_i
    - y_j) \nonumber \\
    &= -c \sum_j l_{ij} \cdot \delta_{ij} \cdot (y_i - y_j)
\end{align}
\vspace*{-0.35cm}

where $l_{ij} = (\mathcal{A}_{ij}$ - $\mathcal{R}_{ij})$ and $\delta_{ij} = \frac{d\; k_y(||y_i - y_j||_2^2)}{d\; ||y_i - y_j||_2^2}$ act as
scalars\footnote{Note that if $k_y$ is a linear kernel, then this term is constant.} on the vector $(y_i - y_j)$ and $\mathcal{A}_{ij}$ and $\mathcal{R}_{ij}$
are functions of $k_x(x_i, x_j)$ and $k_y(y_i, y_j)$.  In this sense, we have that $y_i$ is attracted to $y_j$ according to $\mathcal{A}_{ij} \cdot \delta_{ij}$
and repelled according to $\mathcal{R}_{ij} \cdot \delta_{ij}$. For UMAP, assuming the hyperparameters are set as in~\cite{actup}, the attractions and
repulsions respectively are:
\begin{align}
    \label{eq:umap_grads}
    \nabla_{y_i}^{UMAP} &= \sum_{j, j \neq i} \frac{k_x(x_i, x_j)}{k_y(y_i, y_j)} \delta_{ij} (y_i - y_j) \nonumber \\
                        & -\sum_{k, k \neq i} \frac{1 - k_x(x_i, x_j)}{1 - k_y(y_i, y_j)} \delta_{ij} (y_i - y_j),
\end{align}

where $\delta_{ij} = -2k_y(y_i, y_j)^2$ for $k_y(y_i, y_j) = (1 + ||y_i - y_j||_2^2)^{-1}$. Gradient descent then entails applying these attractions and
repulsions across all $n^2$ pairs of points. 

\paragraph{Generalization}

This attraction/repulsion framework is not restricted to the gradient of the KL-divergence. Since $||y_i - y_j||_2^2 = y_i^\top y_i - 2 y_i^\top y_j + y_j^\top
y_j$, its gradient with respect to $y_i$ immediately gives us the vector between points: $\frac{\partial \; ||y_i - y_j||_2^2}{\partial \; y_i} = 2(y_i - y_j)$.

By the chain rule, then, the gradient of $k_y(||y_i - y_j||_2^2)$ will necessarily be some scalar acting on $(y_i - y_j)$.  Furthermore, any loss function of
the following form will incur gradients with attractions and/or repulsions between points:

\vspace*{-0.5cm}
\begin{equation}
    \label{eq:general_loss_function}
    \mathcal{L}^{\mathcal{F}}(\X, \Y) = \sum^n_{\substack{i, j = 1 \\ i \neq j}} \mathcal{F}\left(k_x(x_i, x_j), k_y(||y_i - y_j||_2^2)\right)
\end{equation}
\vspace*{-0.35cm}

We will use the term \emph{attraction/repulsion dimensionality reduction framework} (ARDR framework) to refer to a DR analysis that uses attractions and
repulsions. We will refer to the task of finding an optimal such rank-$d$ embedding as the \emph{ARDR objective}. That is, $\argmin_Y
\mathcal{L}^{\mathcal{F}}(\X, \Y).$

\subsubsection{In practice}
\label{sssec:practice}

Many gradient-based DR methods develop the above framework before delving into heuristics that are quicker to calculate. For example, a common trick is to
notice that an exponential kernel for $k_x$ is likely to be 0 for distant points. Since $\mathcal{A}_{ij}$ depends linearly on $k_x(x_i, x_j)$ in most ARDR
methods, the attractions are therefore weak if $x_i$ and $x_j$ are far away. To this end, the common heuristic is to compute a $k$-NN graph over \X and only
calculate attractions along pairs of nearest neighbors.

Thus, the usual optimization strategy is to simulate gradient descent by iterating between the relevant attractions and sampled repulsions acting on each point.
Specifically, one attracts $y_i$ to the points corresponding to $x_i$'s $k$ nearest neighbors in \X. Then one chooses $ck$ points from which to repel $y_i$
(either exactly or approximately), where $c$ is some appropriately chosen constant.  \cite{umap_effective_loss} showed that these heuristics
significantly affect the optimization objective, causing the practical implementation to diverge from the stated gradients in Eq.~\ref{eq:umap_grads}.

\subsection{Related Work}
\label{ssec:related_work}

As the popularity of ARDR methods has grown, so too has the literature that disputes the common wisdom surrounding them. We recommend~\cite{tsne_effectively}
to increase intuition regarding what information the tSNE embedding gives regarding its input. We note that UMAP has similar issues to those raised for tSNE.
Correspondingly, deeper analysis into ARDR methods seems to raise more questions than answers. For example, \cite{attr_rep_spectrum} showed that UMAP's
seemingly stronger attractions than tSNE's are a result of the sampling strategy rather than the topological properties of the algorithms. Thus, in accordance
with \cite{umap_effective_loss}, we will refer to UMAP's \emph{effective} objective as the output given by Section \ref{sssec:practice} and its \emph{intended}
objective as the one defined in Equation \ref{eq:general_loss_function} under the KL-divergence. Lastly, the manifold learning intuitions regarding ARDR methods
were questioned by~\cite{tsne_umap_init,trimap,dlp}, where it was shown that it is unclear whether tSNE and UMAP are preserving local/global structure in the
expected manner.  Further analysis of ARDR loss functions can be found in~\cite{pacmap, min_distortion_emb}.

\begin{figure*}
    \centering
    \includegraphics[width=0.9\linewidth]{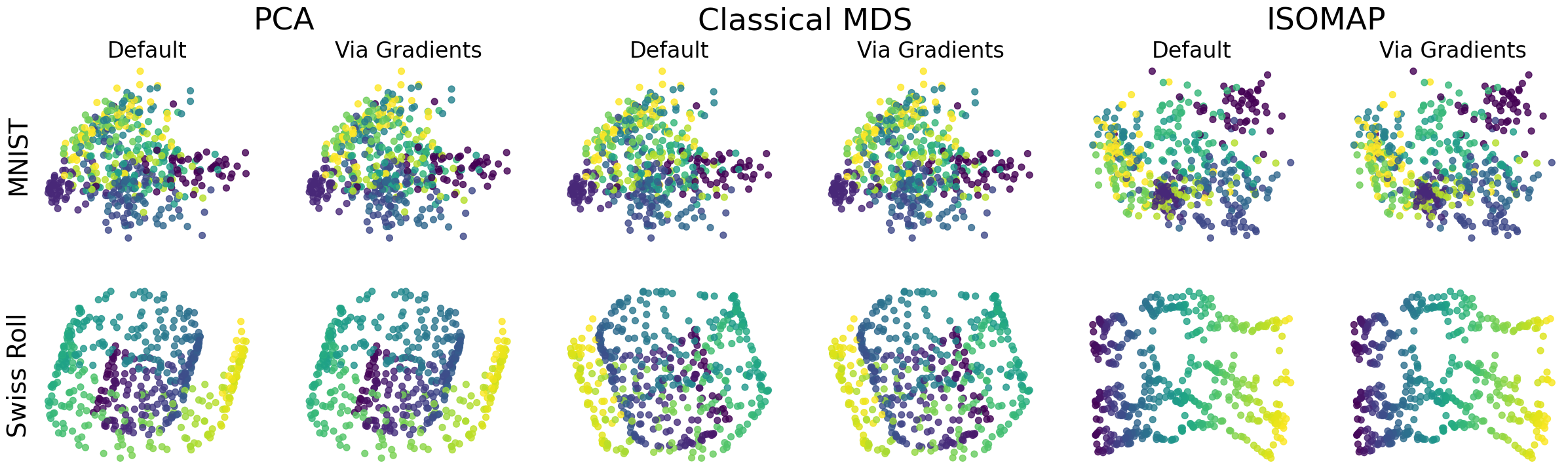}
    \caption{Experimental verification of convergence for PCA, classical MDS and ISOMAP. We show the embeddings for each DR technique
    using the default method and via gradient descent on the points. We use the $L_1$-distance for the Classical MDS setting.}
    \label{fig:theorem_verification}
\end{figure*}

Other works have sought to give insight into DR methods by unifying them. \cite{actup} showed that tSNE and UMAP are effectively identical up to
a hyperparameter and \cite{tsne_umap_contrastive} showed that they can both be reproduced within the contrastive learning framework.  We are less aware of
literature comparing classical DR techniques to the modern wave of gradient-based ones. \cite{min_distortion_emb} defines a framework to unify classical and
modern DR methods, but the generality of the approach makes it difficult to make direct claims regarding the connection between methods like PCA and UMAP.

Lastly, ~\cite{localexp} and ~\cite{dlp} provide explainability tools for DR methods. In the former, the authors use interpretable methods such as PCA to
create local explanations for a given black-box DR method (such as tSNE). In the latter the authors analyze the manifold-learning properties of popular DR
approaches towards the goal of improved explainability. They define a novel manifold-preservation measure and show that (1) tSNE/UMAP preserve locality better
than most classical methods\footnote{Although there are situations where PCA performs best.}, and (2) that tSNE/UMAP distort the Euclidean relationships more
strongly than other DR methods. We are able to reproduce these results with a similar metric in Section~\ref{sec:futurework}. We share the authors' surprise
regarding the lack of literature on explaining DR embeddings since dimensionality reduction is a natural step when gaining intuition regarding distributions of
points.

\section{Classical Methods in the ARDR Framework}
\label{sec:pca_ar}

Representing a learning problem in the ARDR framework requires two things. First, we must find the $k_x$, $k_y$ and $\mathcal{F}$ such that the objective can be
written in the form of Equation \ref{eq:general_loss_function}. Second, we must show that the minimum of the ARDR formulation is indeed the optimum of the
original problem.

\subsection{PCA as Attractions and Repulsions}
\label{ssec:pca_in_ar}

PCA is often presented as an SVD-based algorithm to find the optimal low-rank representation of $\C\X$. By the Eckart-Young-Mirsky theorem, the optimal low-rank
representation with respect to the Frobenius norm is given by the $\tilde{\X}$ s.t. $||\C\X - \tilde{\X}||_F$ is minimized subject to $\text{rank}(\tilde{\X})
\leq d$. It is well-known that, for centered \X, this is given by the first $d$ principal components of $\C \Gx \C = \C \X \X^\top \C$, which is exactly the
embedding \Y obtained by PCA.  With this as justification, we state PCA's objective function as \[ \min_{\Y} ||\C(\Gx - \Gy) \C||_F^2 \] where $\Gy = \Y
\Y^\top$. Note that the centering matrix \C is symmetric and idempotent, i.e. $\C^\top \C = \C^2 = \C$. Thus, we have that $k_x$ and $k_y$ are both the standard
inner product and $\mathcal{F}(a, b) = (a - b)^2$. We now show that this is minimized if and only if \Y is the PCA projection of \X.

\begin{lemma}
    \label{lma:pca_minimum}
    The minimum of $\mathcal{L}^{PCA}(\X, \Y) = || \C ( \Gx - \Gy) \C ||_F^2$ is only obtained when $\Y$ is the PCA projection of $\X$ up to orthogonal transformation.
\end{lemma}

The proof is given in section \ref{pca_lemma_proof} of the supplementary material. We now provide the formula for the PCA gradient and verify that it can be
represented as attractions and repulsions between points.

\begin{corollary}
The PCA gradient $\nabla_{PCA} \in \mathbb{R}^{n \times d}$ is
\begin{equation}
    \label{pca_grad_formula}
    \nabla^{PCA} = -4 \C(\Gx - \Gy)\C\Y.
\end{equation}
\vspace*{-0.4cm}
\end{corollary}

We derive this in Section \ref{pca_grad_derivation} of the supplementary material. Furthermore, the next lemma (proof given in Section \ref{apx:pca_ar}) states
that this gradient can be written in terms of attractions and repulsions.

\begin{lemma}
    \label{lma:pca_ARDR}

    Let $\Lmat \in \mathbb{R}^{n \times n}$ be any matrix of the form $\Lmat = \C \A \C$ for $\A\in \mathbb{R}^{n \times n}$ and let $\alpha$ be a constant.
    Then any gradient of the form\footnote{The $A:dB$ operation represents the Frobenius inner product between matrix $A$ and the differential of matrix $B$.}
    $\nabla = \alpha \Lmat : d\left( \C \Gy \C \right)$ can be expressed via attractions and repulsions on \Y.  Furthermore, the ARDR gradient acting on $y_i$
    has the form $\nabla_{y_i} = c \sum_j l_{ij} (y_i - y_j)$.

\end{lemma}

We verify this experimentally in Figure \ref{fig:theorem_verification}, where we show that the embeddings obtained by gradient descent are equivalent to the
default ones. In essence, one can think of the gradient updates as analogous to the power-iteration method for finding eigenvectors.

\subsection{PCA Convergence} 
\label{sec:pca_convergence}

%
%
%
%
%

The proof is due to the fact that Classical MDS can be performed using PCA.  We are thus guaranteed to obtain the PCA embedding of \X or Classical MDS embedding
of \Dx if we randomly initialize the pointset $\Y$ and apply the corresponding gradient updates.

Performing this gradient descent is impractical, however, as one must calculate a new matrix product at every epoch. In many cases, however, our similarity
functions are psd and are therefore suitable for fast approximations. Given psd matrix \A and its SVD-based rank-$k$ approximation $\A^k$, there are
sublinear-time methods~\cite{sublinear_approx} for obtaining $\A'$ such that 

\vspace*{-0.35cm}
\begin{equation}
    \label{eq:approx_gram}
    ||\A - \A'||_F^2 \in (1 \pm \varepsilon) ||\A - \A^k||_F^2
\end{equation}
\vspace*{-0.35cm}

Our next result shows that these approximations do not significantly affect the gradient descent convergence. Let $\nabla$ be
the full PCA gradient and $\nabla'$ be the gradient obtained using the approximations in Eq.~\ref{eq:approx_gram}. 

\begin{theorem}
    \label{grad_inner_product_lma}
    Let $\lambda_{x_i}$ be the i-th eigenvalue of \Gx and let $\Gx^k$ be the optimal rank-$k$ approximation of \Gx. Then $\langle \nabla, \nabla' \rangle_F > 0$
    as long as $||\Gx - \Gy||_F^2 \geq (1 + \varepsilon) \dfrac{\lambda_{x_1}}{\lambda_{x_k}} ||\Gx - \Gx^k||_F^2.$
\end{theorem}

We prove this in Section \ref{grad_inner_product_proof} of the supplementary material. This effectively means that, as long as \Gy is a worse\footnote{Up to
scaling by $\varepsilon$ and the eigengap} approximation to \Gx than $\Gx^k$, the approximate gradient points in a similar direction to the true one. 
Once this is not the case, we have that $ ||\Gx - \Gy||_F^2$ is an $\alpha$-approximation of $||\Gx - \Gx^k||_F^2$, with $\alpha = (1 + \epsilon)
\frac{\lambda_{x_1}}{\lambda_{x_k}}$. We conjecture that this could lead to a fast method for approximating PCA that does not require an
eigendecomposition -- simply initialize a \Y and perform the gradient updates in sublinear time.

\section{Reproducing UMAP using Classical Methods}
\label{sec:two_kernels}

Using the blueprint developed in sections~\ref{sec:prelim} and~\ref{sec:pca_ar}, we now show that ARDR methods such as UMAP can be emulated using classical
methods and gradient descent. The key observation is simply that ARDR methods have a kernel on \Y while classical methods do not. Unfortunately, this kernel is
also what introduces many of the difficulties when analyzing convergence properties. We will use the prefix DK (double-kernel) when describing a classical
method with two non-linear kernels.

\subsection{PCA with Two Kernels (DK-PCA)}
\label{ssec:pca_2_kernel}

The DK-PCA objective is given by 
\begin{equation}
    \mathcal{L}^{DK-PCA}(\X, \Y) = ||\Kx - \Ky||_F^2,
\end{equation}
where we assume \Kx and \Ky are double-centered.

Consider that Section \ref{ssec:pca_in_ar}'s results relied on a linear (and centered) kernel on \Y. We used this in the proofs by noting that, for linear
\Ky, the $\Kx - \Ky$ term is $0$ along the first $d$ components (since \Y has rank $d$).  Thus, $(\Kx - \Ky) \Y = 0$ in Eq. \ref{pca_grad_formula} by orthogonality
when \Y is the PCA embedding. However, a non-linear $k_y$ impedes this line of reasoning as we may have $\text{rank}(\Ky) > d$. Furthermore, the DK-PCA
objective necessitates a non-constant $\delta_{ij}$ term in its gradient:

\vspace*{-0.4cm}
\begin{equation}
    \label{eq:nonlinear_pca_grad}
    \nabla_{y_i}^{DK-PCA} = -c' \sum_j \left[ \Kx - \Ky \right]_{ij} \delta_{ij} (y_i - y_j).
\end{equation}
\vspace*{-0.4cm}

\begin{figure}
    \centering
    \includegraphics[width=0.85\linewidth]{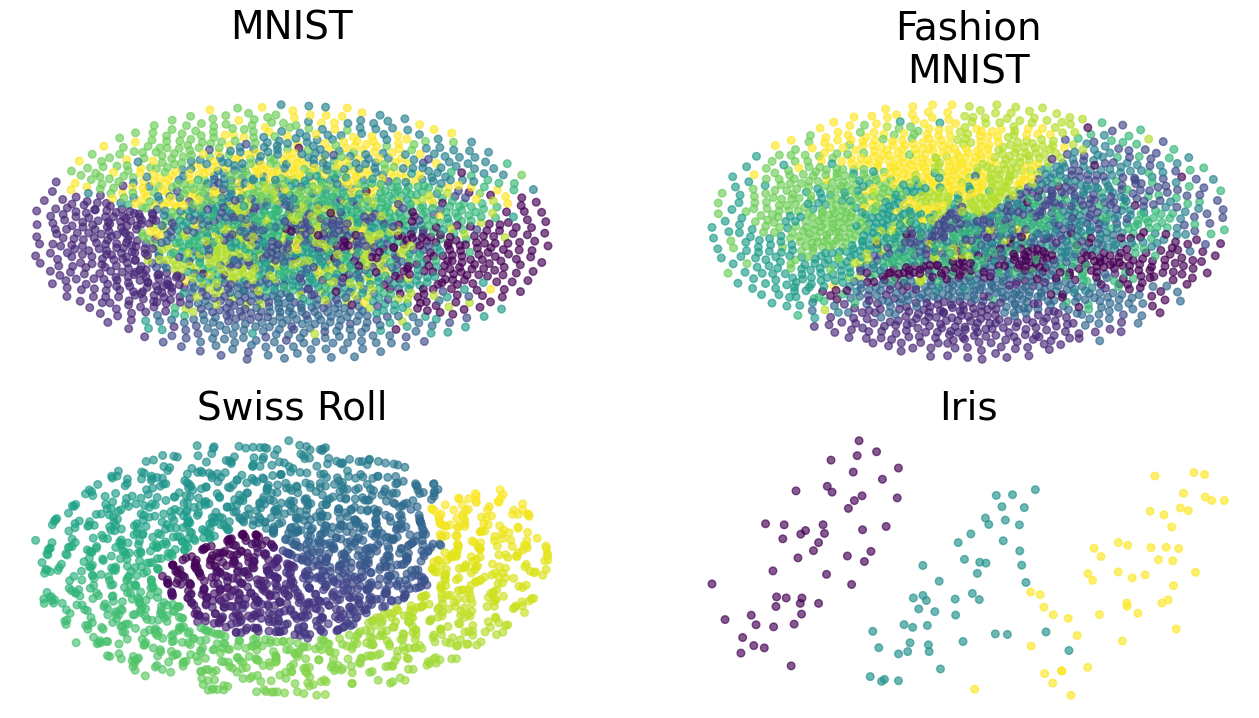}
    \caption{DK-PCA embeddings on the MNIST, Fashion-MNIST, Swiss-Roll and Iris datasets.}
    \label{fig:dkpca_embeddings}
\end{figure}

It was shown in \cite{actup} that applying Eq.~\ref{eq:nonlinear_pca_grad}'s gradients with UMAP's effective optimization scheme provides embeddings that are
very similar to the standard UMAP embeddings. This is verified in column 1 of Figure \ref{fig:dk_embeddings}, where we see that applying UMAP's heuristic
optimization to the DK-PCA objective produces embeddings with similar structure and $k$-NN accuracy to standard UMAP embeddings. However, we show in Figure
\ref{fig:dkpca_embeddings} that this does not hold when optimizing all $n^2$ DK-PCA terms -- the extra repulsions encourage the embedding to spread out over the
available space. In essence, the heuristics in \ref{sssec:practice} significantly alter the embedding and must be present when recreating ARDR methods using
classical techniques.

\subsection{LLE with Two Kernels (DK-LLE)}
\label{sec:lle_in_ar}

Given this motivation, double-kernel LLE (DK-LLE) seems the natural choice to reproduce ARDR methods via classical techniques.  Recall that step 2 of LLE
finds the \Y that best approximates local neighborhoods in \X. The corresponding DK-LLE reasoning is that we want the \Y such that \Ky's local
neighborhoods represent the local neighborhoods in \Kx. To this end, we assume that the LLE covariance matrix constraint $\frac{1}{n} \Y^\top \Y = \I$
should apply in kernel space. The closest substitute is $\frac{1}{n} \Ky = \I$, as $\psi(\Y)$ is then an orthogonal basis and will have a diagonal
covariance matrix. Given \W and \M as in Section \ref{ssec:classical}, the DK-LLE objective is:

\vspace*{-0.35cm}
\begin{equation}
    \label{eq:dklle_loss}
    \mathcal{L}^{DK-LLE}(\W, \Y) = \tr (\M \Ky) + \frac{1}{n}\sum_{i, j}k_y(||y_i - y_j||)_2^2.
\end{equation}
\vspace*{-0.35cm}

\begin{figure}
    \centering
    \begin{subfigure}[b]{\textwidth}
        \includegraphics[width=0.45\textwidth]{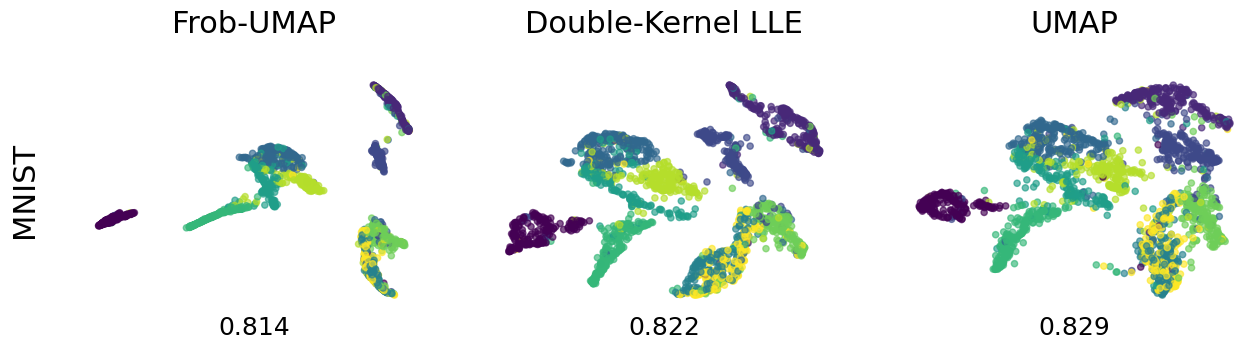}
    \end{subfigure}
    \begin{subfigure}[b]{\textwidth}
        \includegraphics[width=0.45\textwidth]{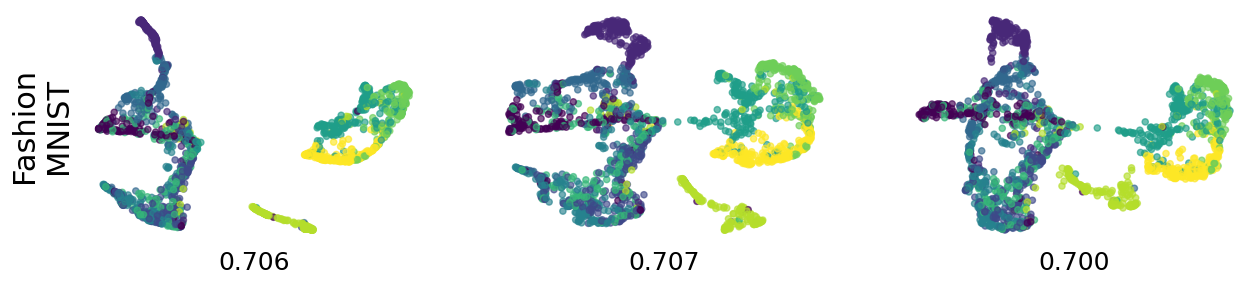}
    \end{subfigure}
    \begin{subfigure}[b]{\textwidth}
        \includegraphics[width=0.45\textwidth]{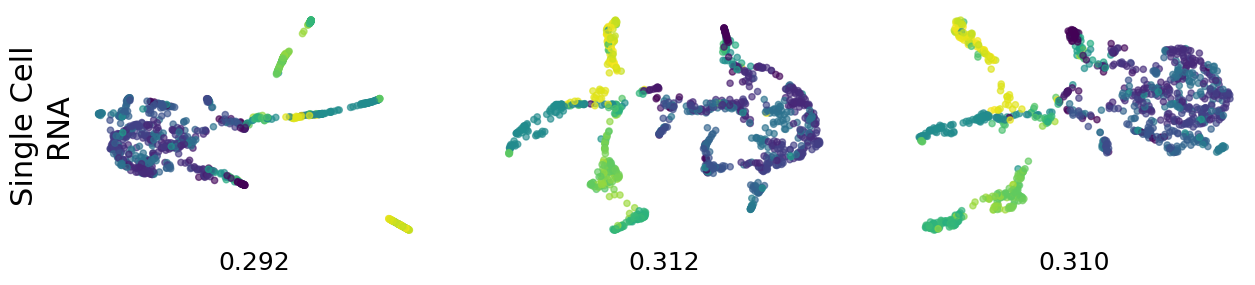}
    \end{subfigure}
    \caption{Embeddings under double-kernel optimization paradigms. kNN-classifier accuracy listed under each plot.}
    \label{fig:dk_embeddings}
\end{figure}

Given this loss function, assume that we have the \W that represents nearest neighborhoods in \X under $k_x$. We now want to use gradient descent to find the \Y
such that $\mathcal{L}^{DK-LLE}(\W, \Y)$ is minimized. This has a natural interpretation in the ARDR framework:

\begin{proposition}
    \label{prop:AR_lle}

    Let \Y and \M be defined as above, $\mathbf{V} = \W^\top \W$, and $c$ be a constant. Then the gradient of Eq. \ref{eq:dklle_loss} with respect to $y_i$ is
    given by \[ \nabla_{y_i}^{DK-LLE} = c \sum_{j} \left(w_{ij} + w_{ji} - v_{ij} - \frac{1}{n}\right) \delta_{ij} (y_i - y_j).\]

\end{proposition}
\noindent The proof can be found in Section \ref{apx:LLE_grad} of the appendix.

\begin{figure}
    \centering
    \includegraphics[width=0.8\linewidth]{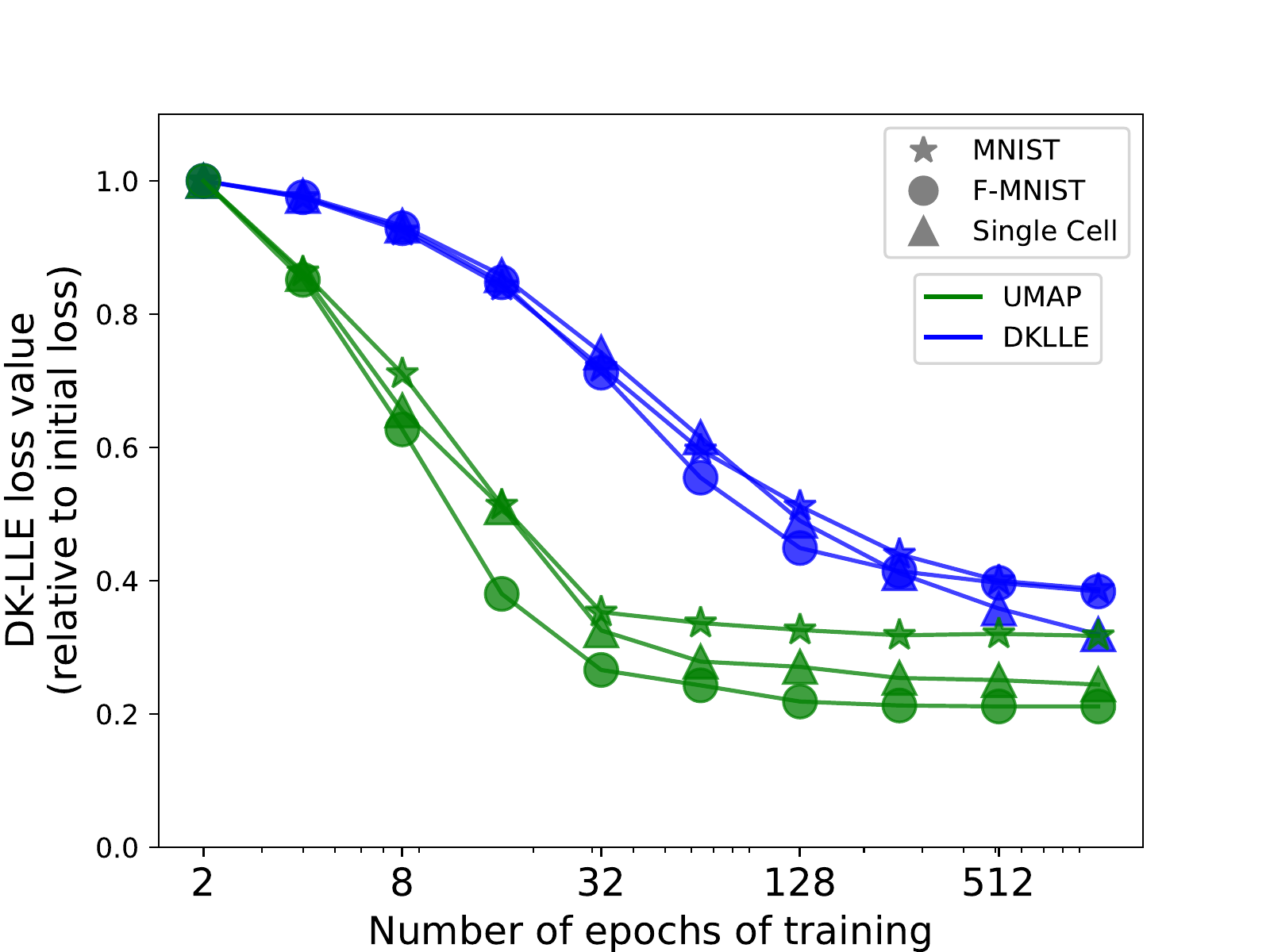}
    \caption{Normalized values of Eq. (\ref{eq:dklle_loss}) as we optimize via UMAP (green) and standard gradient descent (blue). Both methods use the same learning
    rate.}
    \label{fig:dklle_loss}
\end{figure}

\paragraph{Similarities to UMAP.} We take a moment to consider the striking similarities between this and UMAP's effective gradients. Note that the
$\delta_{ij}(y_i - y_j)$ terms are identical between all double-kernel methods and only depend on the choice of $k_y$. As a result, the difference between UMAP
and DK-LLE can only be in the scalars applied onto them.  For $k_y \sim (1 - ||y_i - y_j||_2^2)^{-1}$ as in UMAP, the $\delta_{ij}$ term is always negative.
Given this, the terms $w_{ij}$ and $w_{ji}$ constitute our attraction scalars, i.e. the attraction between $y_i$ and $y_j$ depends exclusively on how much $x_j$
(resp. $x_i$) contributes to $x_i$'s (resp. $x_j$'s) local neighborhood under $k_x$. This is fully in line with the UMAP attraction scalars that are applied
along nearest neighbors in \X.

Now consider the repulsive terms in Lemma~\ref{prop:AR_lle} given by $v_{ij} + \frac{1}{n} = (n v_{ij} + 1) / n$. In \cite{umap_effective_loss}, the authors
find that UMAP's effective optimization scales the repulsions acting on point $y_i$ by $d_i = \sum_j k_x(x_i, x_j)$, which is only calculated over the $k$
nearest neighbors. That is, the strength of the repulsion between $y_i$ and $y_j$ depends on the local neighborhoods of $x_i$ and $x_j$. This is
\emph{precisely} what is being represented in Prop.  \ref{prop:AR_lle}. Consider that, since $\W$ is an adjacency matrix over the $k$-nn graph, the $v_{ij}$
term is the sum of weights over all length-2 paths from $x_i$ to $x_j$ under the kernel $k_x$. In this sense, DK-LLE's repulsions are fully in line with UMAP's
effective objective as defined in \cite{umap_effective_loss}.


Thus, DK-LLE and UMAP gradients share an abundance of similarities. We visualize the effect this has on the embeddings in Figure~\ref{fig:dk_embeddings}, where
we see that the DK-LLE and UMAP embeddings are both quantitatively and qualitatively indistinguishable from one another. We discuss the datasets and experiment
settings in Section~\ref{supp:datasets} of the Appendix. It's important to note that, for the DK-LLE embeddings, we are simply calculating the value of
Eq.~\ref{eq:dklle_loss} in Pytorch and running standard gradient descent using its autograd functionality. As a result, we use \emph{none} of the UMAP
heuristics that are described in Section \ref{sssec:practice}. For completeness, we include additional datasets in Figure~\ref{fig:other_datasets} in the
Appendix. The results there are fully consistent with those presented here.

This equivalence between UMAP and the DK-LLE objective is further evidenced in Figure \ref{fig:dklle_loss}, which shows the normalized DK-LLE loss over the
course of UMAP and DK-LLE optimizations. Specifically, we see that UMAP obtains quicker convergence on the DK-LLE objective than when we optimize the objective
directly by gradient descent. In this sense, UMAP embeddings are simply fast approximations to the DK-LLE loss function.

\section{Conclusions and Future Work}
\label{sec:futurework}

\subsection{Towards interpreting UMAP}

Given the theoretical and experimental evaluation above, we now formalize our primary conjecture: \emph{with high probability, UMAP obtains a constant-factor
approximation to the DK-LLE objective.} Specifically, suppose we are given a dataset \X with optimal DK-LLE embedding $\Y^{OPT}$ and we obtain \X's UMAP
embedding $\Y'$. Then we conjecture that there exists an $\alpha$ such that, with high probability, $\mathcal{L}^{DK-LLE}(\W, \Y') < \alpha
\mathcal{L}^{DK-LLE}(\W, \Y^{OPT})$, where \W is the weight matrix corresponding to \X. While we have given theoretical and experimental evidence towards this,
we recognize that it has not been proven formally.  Nonetheless, since DK-LLE naturally reproduces UMAP's heuristics, our results imply that one can study of
DK-LLE and ascribe the results to UMAP outputs directly.

This brings us to what one can say about UMAP inputs given their embeddings. Recall that standard LLE is minimized when the neighborhoods of \Y maximally
recreate neighborhoods in \X. Similarly, DK-LLE must maintain this similarity under the kernels, i.e. \Y's neighborhoods under \Ky should approximate \X's
neighborhoods under \Kx.  This has an impact on non-nearest-neighbors as well since the optimal LLE (resp. DK-LLE) embedding must stitch the local neighborhoods
together.  By the above conjecture, UMAP must preserve these relationships under the \Kx and \Ky kernels just like DK-LLE. Thus, if $y_i$ and $y_j$ are near one
another in a UMAP embedding, it must be the case that $x_i$ and $x_j$ have high overlap in their local neighborhoods (up to the \Kx and \Ky kernels). To
quantify this, we define the neighborhood preservation for $x_i$ as $b(x_i, y_i; k) = 1$ if $x_i$ and $y_i$ have the same $k$-th neighbor and $0$ otherwise. We
define the neighborhood preservation of the dataset as $B(\mathbf{X}, \mathbf{Y}; k) = \frac{1}{n}\sum_i b(x_i, y_i; k)$. Now consider

\vspace*{-0.4cm}
\begin{equation}
    \label{eq:nbhd_preservation}
    \frac{\frac{1}{m - l} \sum_{k=l}^m B(\mathbf{X}, \mathbf{Y}; k)}{B(\mathbf{X}, \mathbf{Y}; 1)}.
\end{equation}
\vspace*{-0.4cm}

\begin{figure}
    \includegraphics[width=\linewidth]{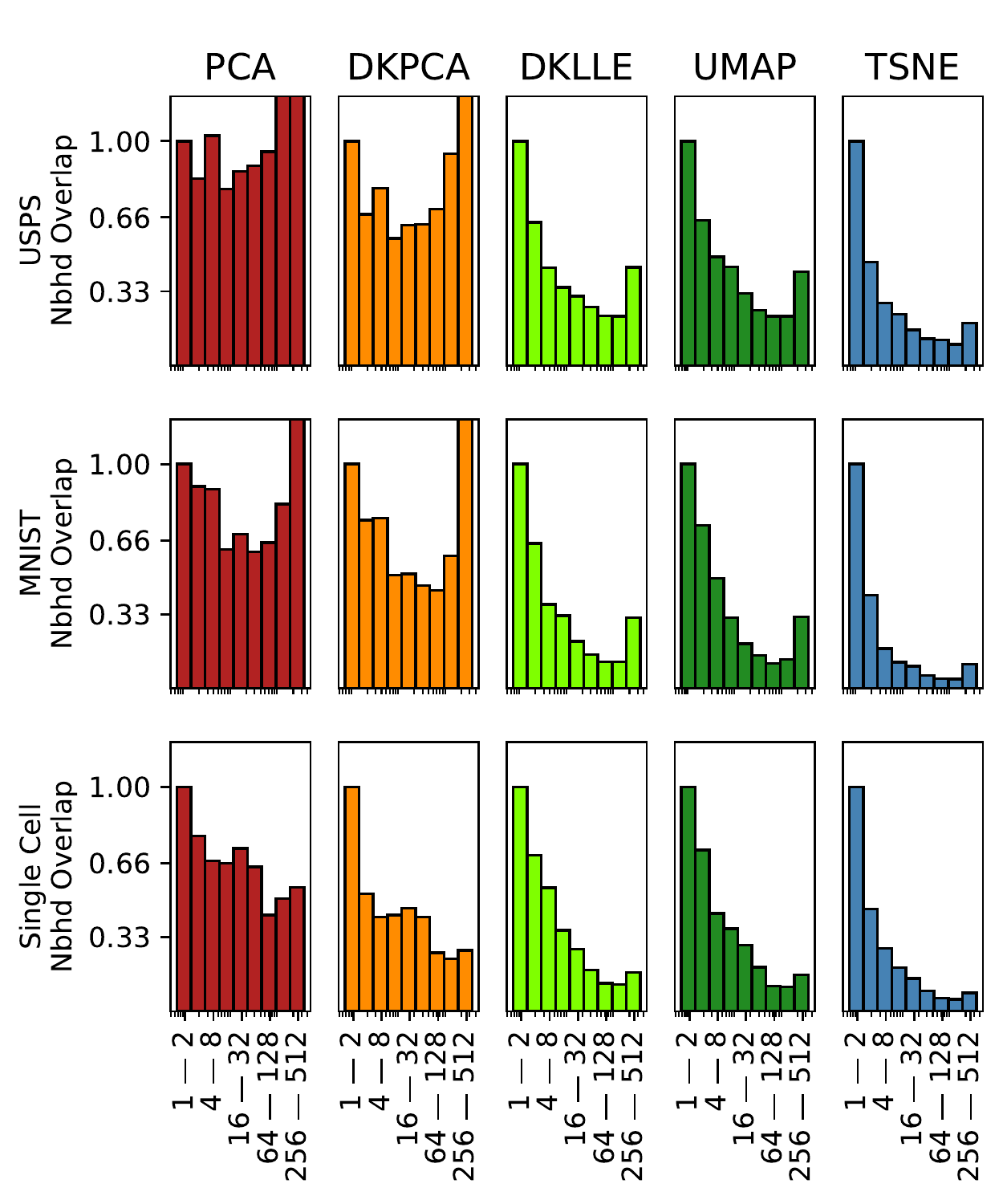}
    \caption{A histogram showing values for Eq. \ref{eq:nbhd_preservation} under various DR algorithms. The $x$-axis shows $l$ and $m$ values.}
    \label{fig:nbhd_preservation}
\end{figure}

This represents the amount that neighbors $l$ through $m$ are preserved relative to how often the nearest neighbor is preserved and is very similar in spirit to
the metrics proposed in \cite{dlp}. We plot these values in Figure~\ref{fig:nbhd_preservation}, which verifies that DK-LLE, UMAP and tSNE have consistent
neighborhood preservation across datasets, and that this preservation degrades consistently as we move from the point in question. Note that UMAP and DK-LLE
have near-identical neighborhood preservations.  Thus, once one knows how much UMAP preserves nearest neighbors, one can state how often the farther neighbors
line up as well. 

\subsection{Future work}

Although we have provided a novel method to analyze UMAP's effective optimization scheme, there remain several open questions. First, it feels natural to ask
how tSNE relates to this. While the relationships between tSNE and UMAP have been widely studied ~\cite{tsne_umap_contrastive, tsne_umap_init, actup, dlp}, we
were unable to recreate tSNE in the DKLLE framework. Furthermore, we believe that the relationship between classical and modern DR methods has a major untouched
component: why can one minimize classical methods such as PCA, ISOMAP, LLE, and Laplacian Eigenmaps via eigendecomposition but the modern ARDR methods all
require gradient descent approximations? It may be that having two kernels makes the problem of finding an optimal embedding NP-hard. If so, does there exist
a way to approximate it without gradient descent?

\printbibliography

\newpage
\appendix
\onecolumn
\section{ARDR Specifics}
\subsection{Examples of tSNE and UMAP being used to draw scientific conclusions}
\label{bad_science_list}

\begin{figure}
    \centering
    \begin{subfigure}[b]{0.24\linewidth}
        \includegraphics[width=\linewidth]{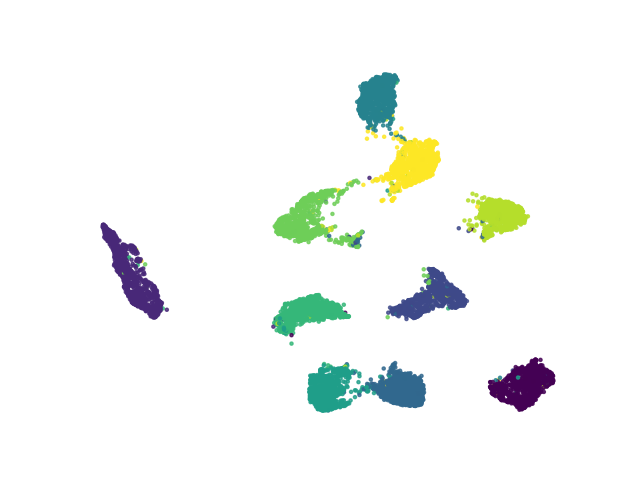}
    \end{subfigure}
    \begin{subfigure}[b]{0.24\linewidth}
        \reflectbox{\includegraphics[width=\linewidth]{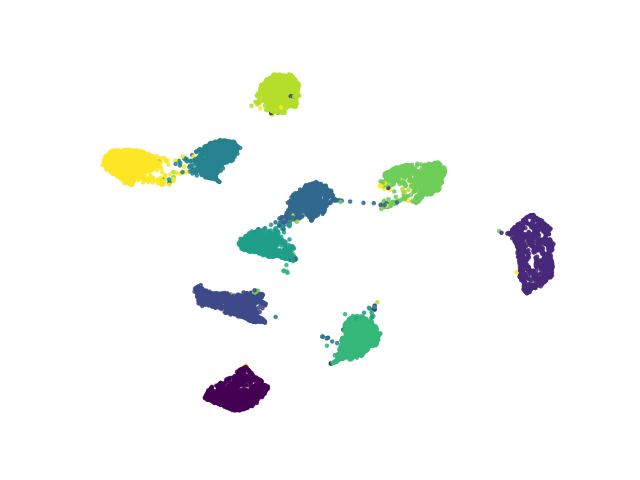}}
    \end{subfigure}
    \begin{subfigure}[b]{0.24\linewidth}
        \includegraphics[width=\linewidth]{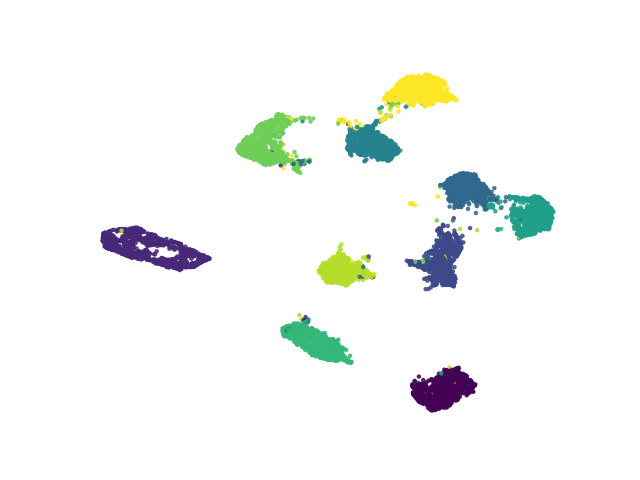}
    \end{subfigure}
    \begin{subfigure}[b]{0.24\linewidth}
        \includegraphics[width=\linewidth]{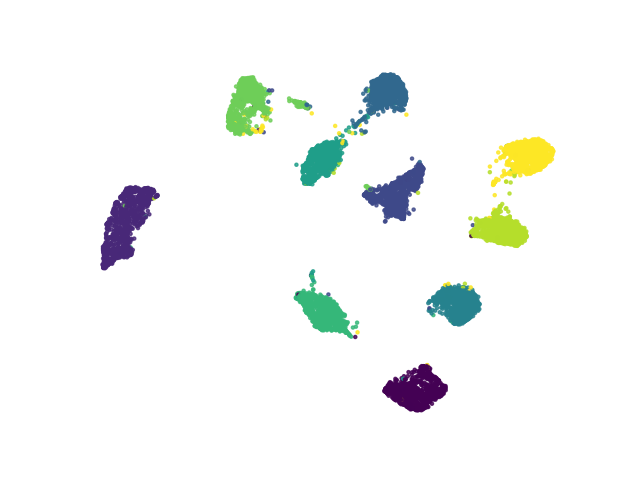}
    \end{subfigure}\\
    
    \begin{subfigure}[b]{0.24\linewidth}
        \reflectbox{\includegraphics[width=\linewidth]{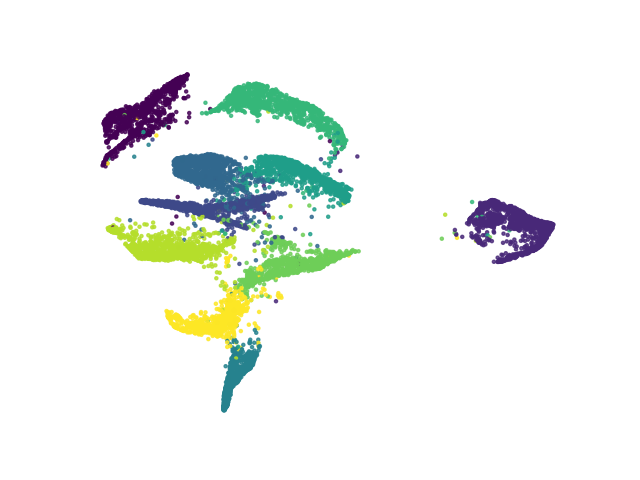}}
    \end{subfigure}
    \begin{subfigure}[b]{0.24\linewidth}
        \includegraphics[width=\linewidth]{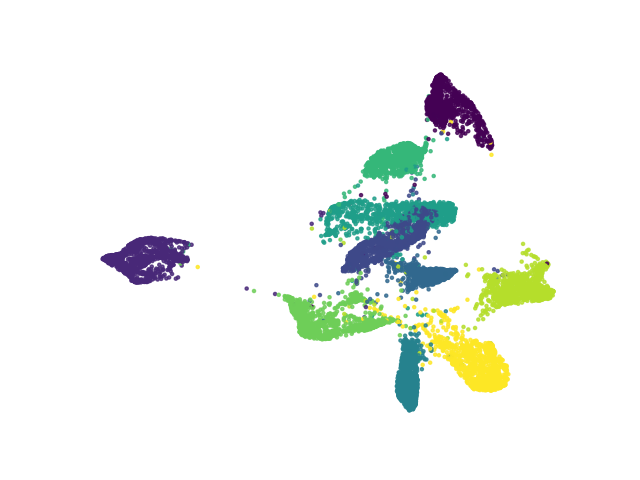}
    \end{subfigure}
    \begin{subfigure}[b]{0.24\linewidth}
        \includegraphics[width=\linewidth]{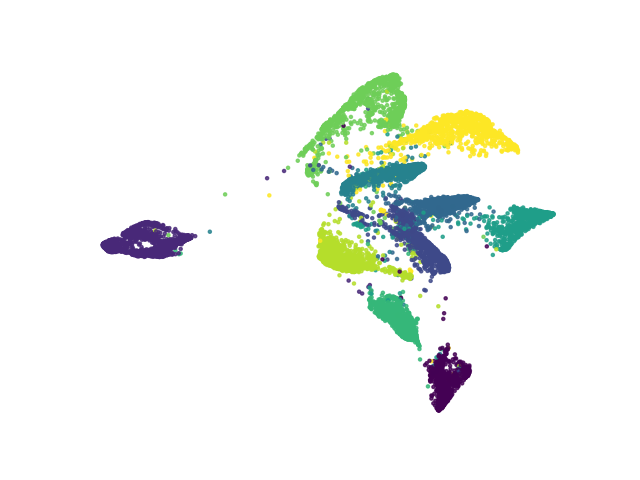}
    \end{subfigure}
    \begin{subfigure}[b]{0.24\linewidth}
        \includegraphics[width=\linewidth]{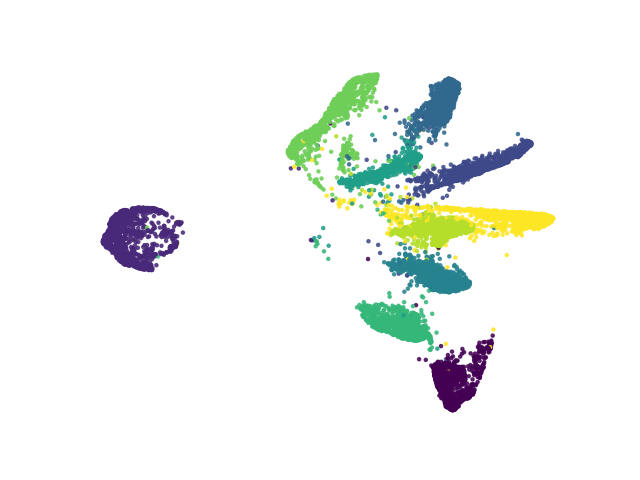}
    \end{subfigure}
    
    \caption{UMAP/tSNE (top/bottom) embeddings of the last conv. layer of VGG11~\cite{vgg} on MNIST mid-training. 3 columns have $>\!\!90\%$ accuracy while one
    has $72\%$ accuracy. We purposefully omit which column corresponds to the low-accuracy latent space.}
    \label{fig:bad_tsne_umap_example}
\end{figure}

We first identify 45 articles that use tSNE or UMAP to embed a neural network's learned latent space: 
\cite{deep_learning_tsne_umap_1,deep_learning_tsne_umap_2,deep_learning_tsne_umap_3,deep_learning_tsne_umap_4,deep_learning_tsne_umap_5,deep_learning_tsne_umap_6,
deep_learning_tsne_umap_7,deep_learning_tsne_umap_8,deep_learning_tsne_umap_9,deep_learning_tsne_umap_10,deep_learning_tsne_umap_11,deep_learning_tsne_umap_12,
deep_learning_tsne_umap_13,deep_learning_tsne_umap_14,deep_learning_tsne_umap_15,deep_learning_tsne_umap_16,deep_learning_tsne_umap_17,deep_learning_tsne_umap_18,
deep_learning_tsne_umap_19,deep_learning_tsne_umap_20,deep_learning_tsne_umap_21,deep_learning_tsne_umap_22,deep_learning_tsne_umap_23,deep_learning_tsne_umap_24,
deep_learning_tsne_umap_25,deep_learning_tsne_umap_26,deep_learning_tsne_umap_27,deep_learning_tsne_umap_28,deep_learning_tsne_umap_29,deep_learning_tsne_umap_30,
deep_learning_tsne_umap_31,deep_learning_tsne_umap_32,deep_learning_tsne_umap_33,deep_learning_tsne_umap_34,deep_learning_tsne_umap_35,deep_learning_tsne_umap_36,
deep_learning_tsne_umap_37,deep_learning_tsne_umap_38,deep_learning_tsne_umap_39,deep_learning_tsne_umap_40,deep_learning_tsne_umap_41,deep_learning_tsne_umap_42,
deep_learning_tsne_umap_43,deep_learning_tsne_umap_44,deep_learning_tsne_umap_45}.

These references use tSNE and UMAP embeddings to verify that the network is extracting the correct features. In essence, if the network has learned different
representations for samples then these samples should be mapped to different locations in the latent space. As a result, the prevailing intuition is that one
can embed this into 2D using tSNE or UMAP and study separation in the latent space by inspecting the distribution in the embedding.

However, we refer to Figure~\ref{fig:bad_tsne_umap_example} as a simple counter-example to this intuition. There, we train a network on MNIST with a low
learning rate and plot the tSNE and UMAP embeddings of the latent space over the course of training. While three of the columns correspond to networks with
$>90\%$ accuracy, one has $\sim \! 70\%$ accuracy. If the intuition held, we should be able to clearly see which column corresponds to a worse learned
representation. We showed this to various coworkers in our department and people had a $\sim \! 50\%$ success rate in identifying which column is the `bad' one.

\section{Further experiments}

\subsection{Experimental procedures}
\label{supp:datasets}

We use the following datasets in our experimental analysis:
\begin{itemize}
    \item MNIST~\cite{mnist}; 60000 images of ten classes with 784 dimensions.
    \item Fashion-MNIST~\cite{fashion_mnist}; 60000 images of ten classes with 784 dimensions.
    \item Single-cell~\cite{single_cell}; 23100 cell descriptors; 152 classes with 45769 dimensions. We use the `cell\_cluster' field as the class label for
        coloring images.
    \item USPS~\cite{usps}; 50000 images of ten classes.
    \item Ionosphere~\cite{ionosphere}: 351 instances of 34 features describing radar data; 2 classes.
    \item Seeds~\cite{seeds}: 210 instances of 7 features describing wheat seeds; 3 classes.
    \item Diabetes~\cite{scikit-learn}: 442 instances of 10 features for diabetes prediction. This is a regression dataset and its labels are values in 25 to
        346..
    \item Abalone~\cite{abalone}: 4177 instances of 8 features for predicting the age of abalone from physical measurements. This is a regression dataset but
        only has $\sim20$ unique values which we use as labels for the $k$-nn classifier in Figure~\ref{fig:other_datasets}.
\end{itemize}

For each dataset, we randomly select $n=1000$ points if it consists of more than $1000$ samples. This is due to runtime concerns for the DK-LLE optimization.

When running DK-LLE, we choose the number of nearest neighbors to be equal to the default in UMAP's base implementation ($k=15$). We initialize via Laplacian
Eigenmaps to stay consistent with UMAP's choice.

\subsection{Remaining datasets}

\begin{figure}
    \centering
    \includegraphics[width=\linewidth]{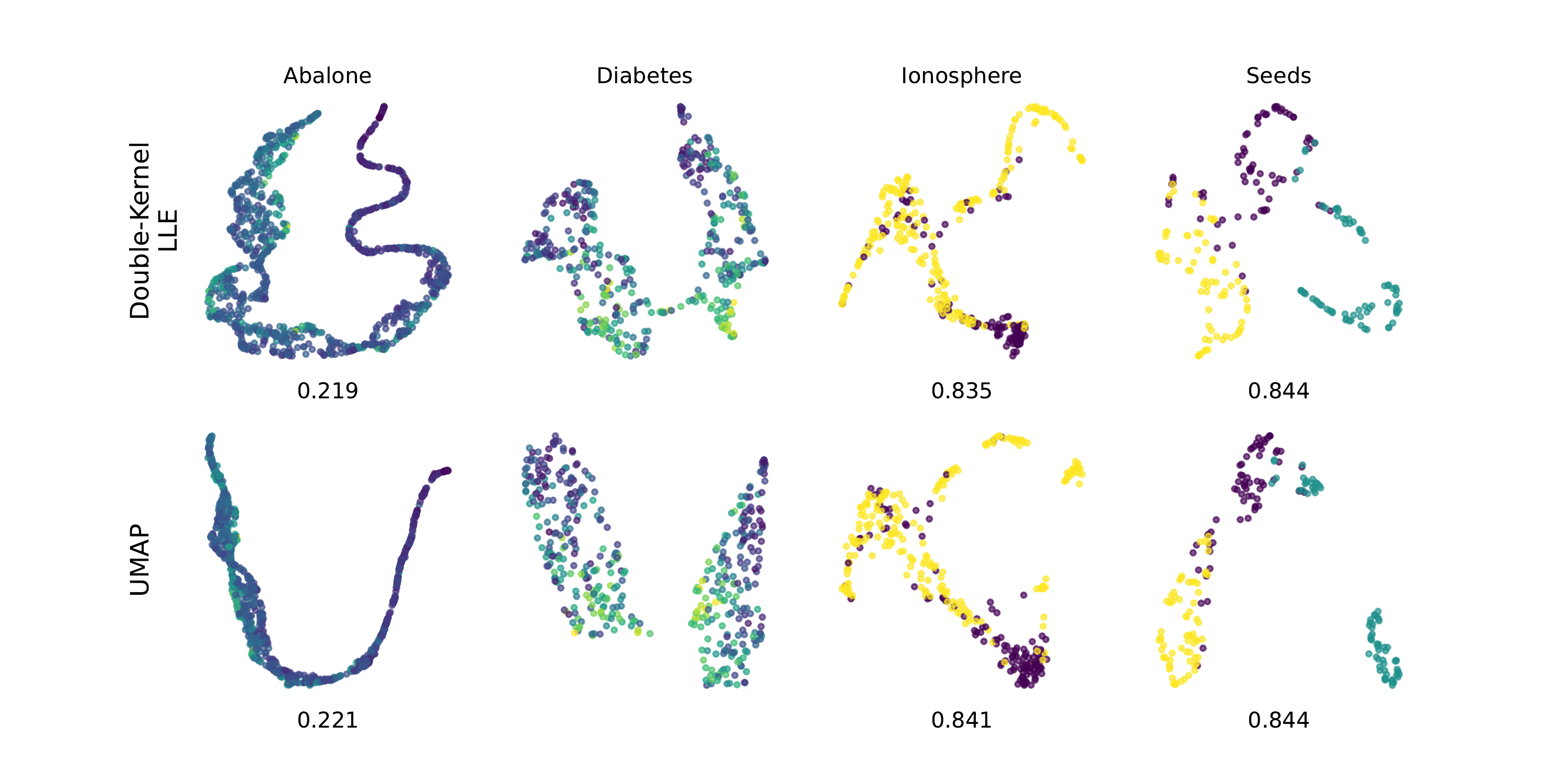}
    \caption{DK-LLE and UMAP comparisons on additional datasets that did not fit in the main body of the paper. The dataset is listed above the top row. $k$-nn
    classifier accuracies are listed below each image. Note, the diabetes dataset is a regression task, implying that $k$-nn classification is not an accurate
    metric.}
    \label{fig:other_datasets}
\end{figure}

For completeness, we plot the DK-LLE and UMAP comparisons on four other datasets in Figure~\ref{fig:other_datasets}. The qualitative and quantitative similarity
between DK-LLE and UMAP remains consistent in every dataset.

\section{PCA theoretical results}

\subsection{Proof of Theorem \ref{lma:pca_minimum}}
\label{pca_lemma_proof}
\begin{proof}
The proof relies on a change of basis in the first $d$ components of $\C\X$ and $\C\Y$.  \label{pca_proof} Let $\C\X = \Ux \Sx \Vx^{\top}$ and $\C\Y = \Uy
\Sy \Vy^{\top}$ be the SVD of $\C\X$ and $\C\Y$. Then $\C\Gx\C = \Ux \Sx^2 \Ux^{\top}$ and $\C\Gy\C = \Uy \Sy^2 \Uy^{\top}$.

We now decompose $\C\X = \Uxp \Sxp (\Vxp)^{\top} + \Uxm \Sxm (\Vxm)^{\top}$, where the $+$ superscript corresponds to the first $d$ components of $\C\X$ and
the $-$ superscript corresponds to the final $D - d$ components. Note that now $\Uxp$ is an orthogonal transformation in the same subspace as $\Uy$. This
allows us to define $\Omat = \Uy (\Uxp)^{\top}$ as the change of basis from $\C\Gx\C$ to $\C\Gy\C$, providing the following characterization of
$L_{PCA}(\X, \Y)$:
\begin{equation}
    \label{pca_change_of_basis}
    L_{PCA}(\X, \Y; \Omat) = ||\Omat \C \Gx \C \Omat^{\top} - \C \Gy \C ||_F^2 = ||(\Uy((\Sxp)^2 - \Sy^2)\Uy^{\top}) + f(\X)||_F^2
\end{equation}
where $f(\X)$ is independent of \Y. The minimum of Equation \ref{pca_change_of_basis} over $\Y$ occurs when $\C\Y$ has the same singular values
as $\Omat\C\X$ in the first $d$ components, which is exactly the PCA projection of $\X$ up to orthonormal transformation.

It remains to show that PCA is invariant to orthonormal transformations. Let $\X' = \Omat \X$. Then $\Gx' = \X'(\X')^{\top} = \Omat \Gx \Omat^{\top} = \Gx$,
since Gram matrices are unique up to orthogonal transformations.
\end{proof}

\subsection{PCA gradient derivation}
\label{pca_grad_derivation}
Let $\Y$ be \textit{any} set of points in $\mathbb{R}^{n \times d}$. Then the gradient with respect to $\Y$ of $\mathcal{L}_{PCA}(\X, \Y) = ||\C\Gx\C
- \C\Gy\C||_F^2$ is obtained by
\begin{align*}
    d \; \mathcal{L}_{PCA}(\X, \Y) &= d \; ||\C (\Gx - \Gy) \C||_F^2 \\
    &= d \; \left( \left( \C(\Gx - \Gy)\C \right)^{\top} : \left( \C(\Gx - \Gy)\C \right) \right)\\
    &= 2 ( \C(\Gx - \Gy)\C )^{\top} : d ( \C(\Gx - \Gy)\C ) \\ &= -2 \C(\Gx - \Gy)\C : d \Gy \\
    &= -4 \C(\Gx - \Gy)\C\Y : d \Y \\
    \frac{d \; f_{PCA}(\X, \Y)}{d\Y} &= -4\C(\Gx - \Gy)\C\Y
\end{align*}
where $:$ represents the Frobenius inner product and the centering matrices cancel due to idempotence.

\subsection{Proof of Lemma \ref{lma:pca_ARDR}}
\label{apx:pca_ar}
\begin{proof}
We use the well-known fact that $\C \Gy \C = -\frac{1}{2} \C \Dy \C$, for squared Euclidean distance matrix \Dy, to express the gradient in terms of $(y_i - y_j)$
vectors.
\begin{align*}
    \nabla &= \alpha \Lmat : d \left( -\frac{1}{2} \C \Dy \C \right)  &\\
    &= -\frac{\alpha}{2} \Lmat : d \Dy \quad & (\text{\C cancels by idempotence})\\ 
    \implies \nabla_{y_i} &= -\alpha \sum_{j} [\Lmat]_{ij} (y_i - y_j) &\\
\end{align*}\end{proof}

\subsection{Proof of Lemma \ref{grad_inner_product_lma}}
\label{grad_inner_product_proof}

The proof relies on defining $\Gx' = \Gx + \Rx$ and $\Gy' = \Gy + \Ry$ as the sum of the original Gram matrices plus residual matrices $\mathbf{R}$. Then
\[  ||\Rx||_F^2 = (1 + \epsilon)||\Gx - \Gx^k||_F^2 = (1 + \epsilon)\sum_{i=k+1}^n \sigma_{xi}^2 \]
where $\sigma_{xi}$ is the i-th singular value of $\Gx$. Notice that since $\Gy^K = \Gy$ for $k \leq d$, we have $\Ry = 0$.

Plugging these in, we get 
\begin{align*}
    \langle \nabla, \nabla' \rangle_F &= \text{Tr}\left[ C (G_X - G_Y) C Y Y^T C (G'_X - G'_Y) C \right] \\
    &= \text{Tr}\left[ C (G_X - G_Y) C G_Y C \left[(G_X - G_Y) - (R_X - R_Y) \right] C \right] \\
    &= \text{Tr}\left[ (G_X - G_Y) C G_Y C \left[(G_X - G_Y) - R_X \right] C \right] \\
    &= \text{Tr}\left[ \left( (G_X - G_Y) C G_Y C (G_X - G_Y) \right) - \left( (G_X - G_Y) C G_Y C R_X \right) C \right] \\
    &= \text{Tr}\left[ (G_X - G_Y)^2 C G_Y C \right] - \text{Tr}\left[ C (G_X - G_Y) C G_Y C R_X \right] \\
\end{align*}
where we perform rearrangements and cancel one of the $C$'s due to the trace being invariant to cyclic permutations.

Notice that the first term $\text{Tr}\left( (G_X - G_Y)^2 C G_Y C \right)$ must be positive as it is the trace of a positive semi-definite matrix. It then
suffices to show that
\[ \text{Tr} \left( (G_X - G_Y)^2 C G_Y C \right) > | \text{Tr} \left( C (G_X - G_Y) C G_Y C R_X \right) | \]
If this condition is satisfied, then we have that $\langle \nabla, \nabla' \rangle_F > 0$, allowing us to employ subgradient descent methods.

\begin{align*}
    \langle \nabla, \nabla' \rangle &\geq \text{Tr}\left( (G_X - G_Y)^2 C G_Y C \right) - | \text{Tr}\left( C (G_X - G_Y) G_Y R_X \right) | \\
    &\geq \text{Tr}\left( (G_X - G_Y)^2 C G_Y C \right) - c \cdot || (G_X - G_Y) (C G_Y C)^{1/2} ||_F \cdot ||(C G_Y C)^{1/2} R_X ||_F \\
    &= || (G_X - G_Y) (C G_Y C)^{1/2} ||_F^2 - c || (G_X - G_Y) (C G_Y C)^{1/2} ||_F \cdot ||(C G_Y C)^{1/2} R_X ||_F \\
    &= || (G_X - G_Y) (C G_Y C)^{1/2} ||_F \left( || (C G_Y C)^{1/2} (G_X - G_Y) ||_F  - c \cdot ||(C G_Y C)^{1/2} R_X ||_F \right)
\end{align*}
where $c = ||C||_F^2$.

Since the first term $|| (G_X - G_Y) (C G_Y C)^{1/2} ||_F$ is necessarily positive, we have that $\langle \nabla, \nabla' \rangle_F \geq 0$ as
long as $|| (C G_Y C)^{1/2} (G_X - G_Y) ||_F - c \cdot ||(C G_Y C)^{1/2} R_X ||_F \geq 0$
We can also remove the $c$ scalar, since we know that $c > 1$. This gives us the necessary condition
\begin{equation}
    \label{grad_nec_condition}
    || (C G_Y C)^{1/2} (G_X - G_Y) ||_F - ||(C G_Y C)^{1/2} R_X ||_F \geq 0 \implies \langle \nabla, \nabla' \rangle_F \geq 0
\end{equation}

We can think of $(C G_Y C)^{1/2}$ as any dataset with the same principal components as $CY$. So this necessary condition is effectively saying that the
inner product between $CY$ and $(G_X - G_Y)$ must be bigger than the inner product between $CY$ and $R_X$. This intuitively makes sense. Consider that $G_X
- G_Y$ is the amount of error in our current projection $CY$. Meanwhile, $R_X$ is an $\epsilon$-approximation of $G_X^k$, the optimal low-rank representation of
$G_X$. So our necessary condition states that as long as $G_Y$ is not an $\epsilon$-approximation of $G_X$, we can continue to use the sublinear-time
approximation of $G_X$ to approximate the gradient $\nabla$.

Said otherwise, if $G_Y$ is sufficiently different from $G_X$, then $\langle \nabla, \nabla' \rangle_F$ is positive. If not, then we have that $G_Y$ is
approximates $G_X^k$, the optimal low-rank approximation of $G_X$. We formalize this below.

Let $G_Y$ be such that $||G_X - G_Y||_F^2 \geq (1 + \alpha) ||G_X - G_X^k||_F^2$ for $\alpha > 0$. Then we want to solve for the $\alpha$ that makes equation
\ref{grad_nec_condition} necessarily positive. This is equivalent to finding the $\alpha$ that satisfies $\min ||(G_X - G_Y) (C G_Y C)^{1/2}||_F = \max ||(C G_X
C)^{1/2} R_X||_F$. We then obtan a lower bound for the minimum:
\begin{align*}
    ||(G_X - G_Y) (C G_Y C)^{1/2}||_F &\geq \sigma_{min} \left( (C G_Y C)^{1/2} \right) \cdot ||G_X - G_Y||_F \\
    &\geq \sigma_{min}\left( (C G_Y C)^{1/2} \right) \cdot \sqrt{ (1 + \alpha)||G_X - G_X^k||_F^2 } \\
\end{align*}
and an upper bound for the maximum:
\begin{align*}
    ||(C G_Y C)^{1/2} R_X||_F &\leq ||(C G_Y C)^{1/2}||_2 ||R_X||_F \\
    &\leq ||(C G_Y C)^{1/2}||_2 \cdot \sqrt{ (1 + \epsilon) ||G_X - G_X^k||_F^2 } \\
    &\leq \sigma_{max}\left((C G_Y C)^{1/2}\right) \cdot \sqrt{ (1 + \epsilon) ||G_X - G_X^k||_F^2 }
\end{align*}

Setting the lower bound equal to the upper bound and solving for $\alpha$ tells us that equation \ref{grad_nec_condition} is greater than 0 when $\alpha$
satisfies
\begin{align*}
    \alpha &> \dfrac{\lambda_{y_1}}{\lambda_{y_k}} \cdot \dfrac{(1 + \epsilon)||G_X - G_X^k||_2^2}{||G_X - G_X^k||_F^2} - 1 \\
    &= (1 + \epsilon) \dfrac{\lambda_{y_1}}{\lambda_{y_k}} - 1
\end{align*}

This means that our gradient $\nabla'$ will be in line with the true gradient $\nabla$ as long as $G_Y$ satisfies the following condition:
\[ ||G_X - G_Y||_F^2 \geq (1 + \epsilon) \dfrac{\lambda_{y_1}}{\lambda_{y_k}} ||G_X - G_X^k||_F^2 \]
As long as this is true, the gradient will push $Y$ into the direction of the PCA embedding of $X$ in $k$ dimensions. If we initialize our $Y$ such that
$\lambda_{y_1} \approx \lambda_{y_k}$, then we know they will slowly diverge over the course of gradient descent. Thus, we can upper bound the ratio
$\lambda_{y_1} / \lambda_{y_k} \leq \lambda_{x_1} / \lambda_{x_k}$. This gives us our final condition for convergence:

\begin{equation}
    \langle \nabla, \nabla' \rangle_F \text{ will be greater than 0 as long as } ||G_X - G_Y||_F^2 \geq (1 + \epsilon) \dfrac{\lambda_{x_1}}{\lambda_{x_k}} ||G_X - G_X^k||_F^2
\end{equation}


\section{LLE Theoretical Results}
\subsection{LLE Derivation}
\label{apx:lle_eig}

If $x_i \approx \sum_{j \in K_i} w_{ij}x_{j}$ is the representation of $x_i$ by a linear combination of its nearest neighbors, then we want to find the \Y such that
$y_i \approx \sum_{j \in K_i} w_{ij} y_{j}$. Treating this as an optimization problem, we can write \[ \min ||\Y - \W \Y||_F^2 \quad \text{s.t.} \quad \Y^\top \Y = \I.
\] By applying a Lagrangian to the constraint and taking the gradient, we have

\begin{align*}
    \min \tr &\left( (\I - \W) \Gy (\I - \W)^\top \right) + \tr\left( \La (\I - \Y^\top \Y) \right)\\
    \Rightarrow \nabla_{\Y} &= 2 \M \Y - 2 \Y^\top \La \\
    \quad \stackrel{\mathclap{\normalfont\mbox{set $\nabla_\Y$ to 0}}}{\Longrightarrow} \quad \quad \M \Y &= \Y^\top \La \\
\end{align*}

where $\M = (\I - \W)^\top (\I - \W)$.  Since we are minimizing the objective, the embedding \Y is given by the eigenvectors of $\M = (\I - \W)^\top (\I - \W)$
that correspond to the smallest $d$ positive eigenvalues as these represent the smallest Lagrangians. Note that $(\I - \W)$ is a graph Laplacian matrix and
therefore \M has at least $1$ zero eigenvalue.

\subsection{Proof of Proposition \ref{prop:AR_lle}}
\label{apx:LLE_grad}
\begin{proof}
We seek the gradient of
\[ \mathcal{L}_{LLE}(\Y) = \tr (\M \Ky) + \frac{1}{n}\tr((\I - \Ky)^2)\]
We start with the first term. By differentiating the Frobenius inner product, we have
\begin{equation}
    \label{eq:inner_prod_LLE}
    \dfrac{\partial \tr (\M \Ky) }{\partial \Y} = \M : \dfrac{\partial \Ky}{\partial \Y}.
\end{equation}
Now consider that the $(i, j)$-th entry of \Ky is a function of $||y_i - y_j||_2^2$ and therefore only depends on the $(i, j)$-th entry of \Dy. Thus,
\[ \dfrac{\partial \Ky}{\partial \Y} = \dfrac{\partial \Ky}{\partial \Dy} \odot \dfrac{\partial \Dy}{\partial \Y}, \]
where $\odot$ is the Hadamard element-wise matrix product. We plug this into Eq. \ref{eq:inner_prod_LLE} and rearrange terms to get
\[ \dfrac{\partial \tr (\M \Ky) }{\partial \Y} = \left[ \dfrac{\partial \Ky}{\partial \Dy} \odot \M \right] : \dfrac{\partial \Dy}{\partial \Y}. \]
We now use the intuition developed in Section \ref{sssec:theory} to point out that this corresponds to the gradient acting on point $y_i$ as
\begin{equation}
    \label{eq:LLE_grad_y_i}
    \nabla_{y_i}(\tr (\M \Ky)) = c \sum_{j} m_{ij} \dfrac{d\; k_y(||y_i - y_j||_2^2)}{d\; ||y_i - y_j||_2^2} (y_i - y_j)
\end{equation}

Now recall that $\M = (\I - \W)^\top (\I - \W) = -\W -\W^\top + \I + \W^\top \W$. Thus, we can write Eq. \ref{eq:LLE_grad_y_i} as
\[ \nabla_{y_i}(\tr (\M \Ky)) = c \sum_{j} (-w_{ij} - w_{ji} + \mathbbm{1}_{i = j} + \left[ \W ^\top \W \right]_{ij}) \dfrac{d\; k_y(||y_i - y_j||_2^2)}{d\; ||y_i - y_j||_2^2} (y_i - y_j), \]
where $\mathbbm{1}_{i=j}$ is 1 if $i=j$ and 0 otherwise. However, notice that if $y_i = y_j$ implies that $y_i - y_j$ is 0. Thus, in the $i=j$ setting the gradient will be 0. Thus, we can cancel the $\mathbbm{1}$ term from the sum, giving the desired result.

It remains to show the gradient of the second term $\frac{1}{n}\sum_{i,j}k_y(||y_i - y_j||_2^2)$. There, it is a simple re-use of the above:
\begin{align*}
    \frac{\partial(\frac{1}{n}\sum_{i,j}k_y(||y_i - y_j||_2^2))}{\partial \Y} &= \frac{1}{n}sum_{i, j}\frac{\partial (k_y(||y_i - y_j||_2^2))}{\partial \Y} \\\
    \implies \nabla_{y_i} &= -\frac{1}{n} \sum_{j} \dfrac{d\; k_y(||y_i - y_j||_2^2)}{d\; ||y_i - y_j||_2^2} (y_i - y_j)
\end{align*}
\end{proof}





\end{document}